\documentclass[10pt,twocolumn,letterpaper]{article}

\usepackage{iccv}
\usepackage{times}
\usepackage{epsfig}
\usepackage{graphicx}
\usepackage{amsmath}
\usepackage{amssymb}

\usepackage{xcolor}
\usepackage{graphicx}
\usepackage{subcaption}
\usepackage{multirow}
\usepackage{microtype}
\usepackage[accsupp]{axessibility}
\usepackage[export]{adjustbox}

\usepackage[breaklinks=true,bookmarks=false]{hyperref}
\iccvfinalcopy 

\def\iccvPaperID{4625} 

\definecolor{todo_color}{rgb}{1.0,0,0.0}
\definecolor{help_color}{rgb}{0.7,0.6,0.2}

\definecolor{mathias_color}{rgb}{.6,.4,.05}
\definecolor{edited_color}{rgb}{.5,.7,.1}
\definecolor{chengcheng_color}{rgb}{0.35,0,0.75}
\definecolor{chris_color}{rgb}{0,0.35,0}
\definecolor{cem_color}{rgb}{0,0,0.85}
\definecolor{markus_color}{rgb}{0,0.35,0.35}
\definecolor{rob_color}{rgb}{0.35,0.35,0}
\definecolor{thomas_color}{rgb}{0.1,0.2,0.8}

\newcommand{\ours}{MotionDeltaCNN\xspace}
\usepackage[capitalize]{cleveref}
\crefname{section}{Sec.}{Secs.}
\Crefname{section}{Section}{Sections}
\Crefname{table}{Table}{Tables}
\crefname{table}{Tab.}{Tabs.}

\makeatletter
\newcommand{\thickhline}{%
    \noalign {\ifnum 0=`}\fi \hrule height 1pt
    \futurelet \reserved@a \@xhline
}
\makeatother

\usepackage{etoolbox}
\newcommand{\insertfig}{\includegraphics[width=\linewidth]{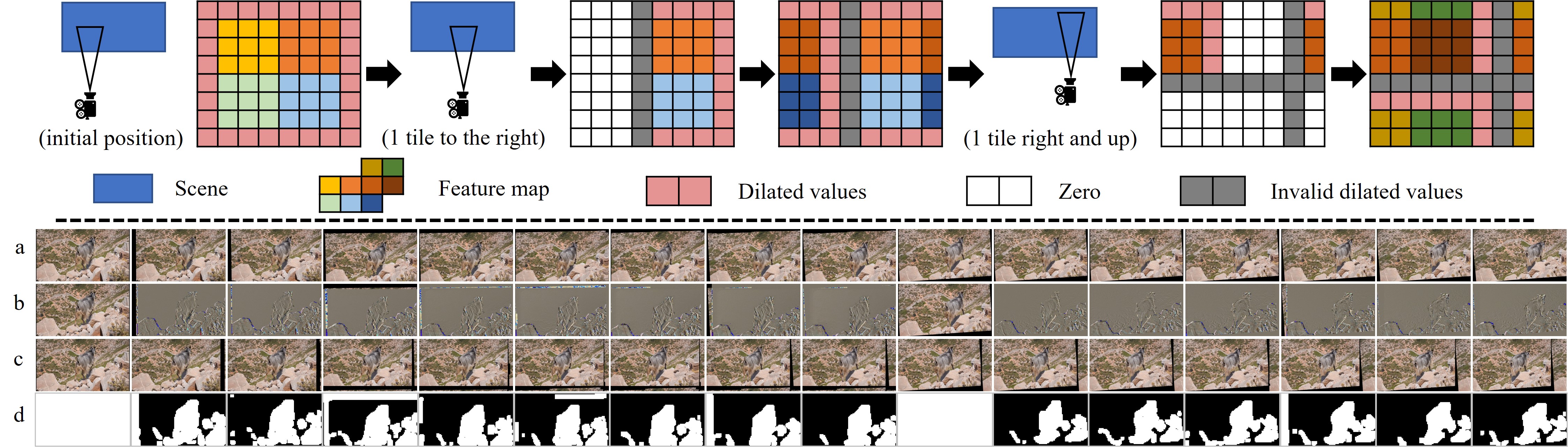}\captionof{figure}{\label{fig:teaser}\ours leverages temporal continuity to accelerate CNN inference for videos with \emph{moving} cameras (bottom) by processing only the sparse frame differences based on a two-dimensional ring buffer, \emph{spherical buffer} (top). Instead of computing each frame individually, we process only the changes (row $b$) between the aligned current (row $a$) and previous frames. Our padded convolution creates dilated values (top) which allow for seamless attachment of newly unveiled tiles onto the existing spherical buffer (row $c$). With these concepts, \ours speeds up inference by processing only new regions or updated pixels (white areas in row $d$) without additional memory allocation.
}\vspace{3pt}}
\makeatletter
\apptocmd{\@maketitle}{\centering\insertfig}{}{}
\makeatother

\begin{document}

\title{MotionDeltaCNN: Sparse CNN Inference of Frame Differences
in Moving Camera Videos with Spherical Buffers and Padded Convolutions}

\author{
Mathias Parger$^{1}$~~~~Chengcheng Tang$^{2}$~~~~Thomas Neff$^{1}$~~~~Christopher D. Twigg$^{2}$\\Cem Keskin$^{2}$~~~~%
Robert Wang$^{2}$~~~~~~Markus Steinberger$^{1}$\\
$^{1}$Graz University of Technology, $^{2}$Meta Reality Labs\\
{\tt\small $^{1}$\{mathias.parger, thomas.neff, steinberger\}@icg.tugraz.at}\\
{\tt\small $^{2}$\{chengcheng.tang, cdtwigg, cemkeskin, rywang\}@meta.com}
}

\maketitle

\begin{abstract}
\vspace{-5pt}
Convolutional neural network inference on video input is computationally expensive and requires high memory bandwidth.
Recently, DeltaCNN\cite{Parger_2022} managed to reduce the cost by only processing pixels with significant updates over the previous frame.
However, DeltaCNN relies on static camera input.
Moving cameras add new challenges in how to fuse newly unveiled image regions with already processed regions efficiently to minimize the update rate - without increasing memory overhead and without knowing the camera extrinsics of future frames.
In this work, we propose \ours, a sparse CNN inference framework that supports moving cameras.
We introduce spherical buffers and padded convolutions to enable seamless fusion of newly unveiled regions and previously processed regions -- without increasing memory footprint.
Our evaluation shows that we outperform DeltaCNN by up to 90\% for moving camera videos.

\end{abstract}

\section{Introduction}

Real-time inference of convolutional neural networks (CNN) with video streams remains a power-consuming task and is often infeasible on mobile devices due to hardware and thermal limitations,
despite recent efforts aiming at efficient CNN inference through pruning \cite{Le1990,Han2015}, quantization \cite{Hubara2018, Moons2018, Lin2016}, specialized hardware \cite{Chen2015, Han2016, Chen2016} or network optimization \cite{Sifre2014,Tan2019} 
Video input allows for performance optimization through temporal similarity between frames.
One common method is to use large, slow networks for accurate predictions at key frames, updated with small, fast networks at intermediate frames \cite{Zhu2017,Zhu2018,Li2018,Jain2019,Fan2020,Liang_2022_CVPR,Habibian2022}.
However, this approach requires special network design and training and is not suitable for significant frame changes.

\begin{figure*}[pt]
\begin{subfigure}{.04\linewidth}
a
\end{subfigure}%
\begin{subfigure}{.31\linewidth}
  \centering
  \includegraphics[width=.95\linewidth,valign=c]{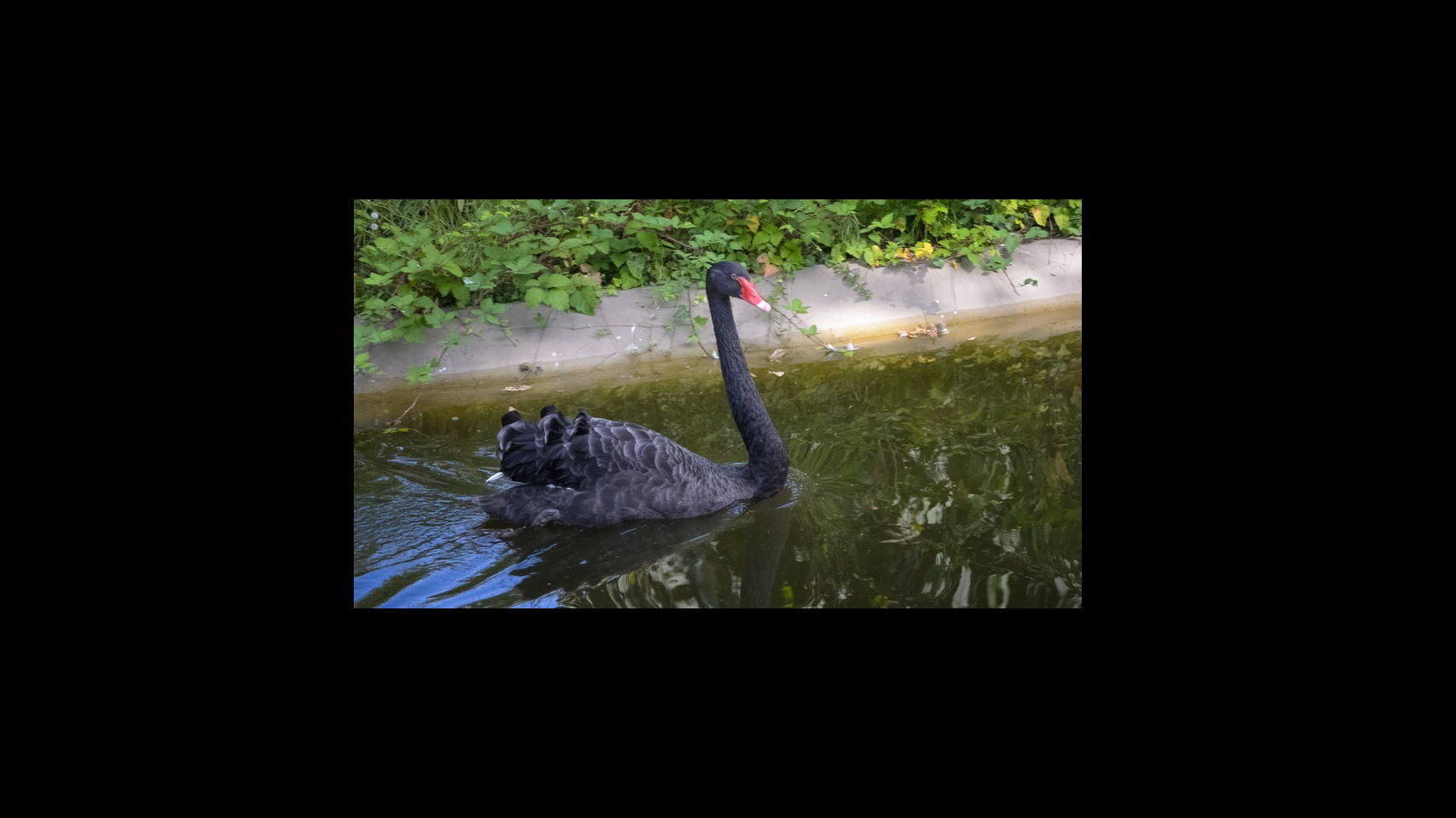}
\end{subfigure}%
\begin{subfigure}{.31\linewidth}
  \centering
  \includegraphics[width=.95\linewidth,valign=c]{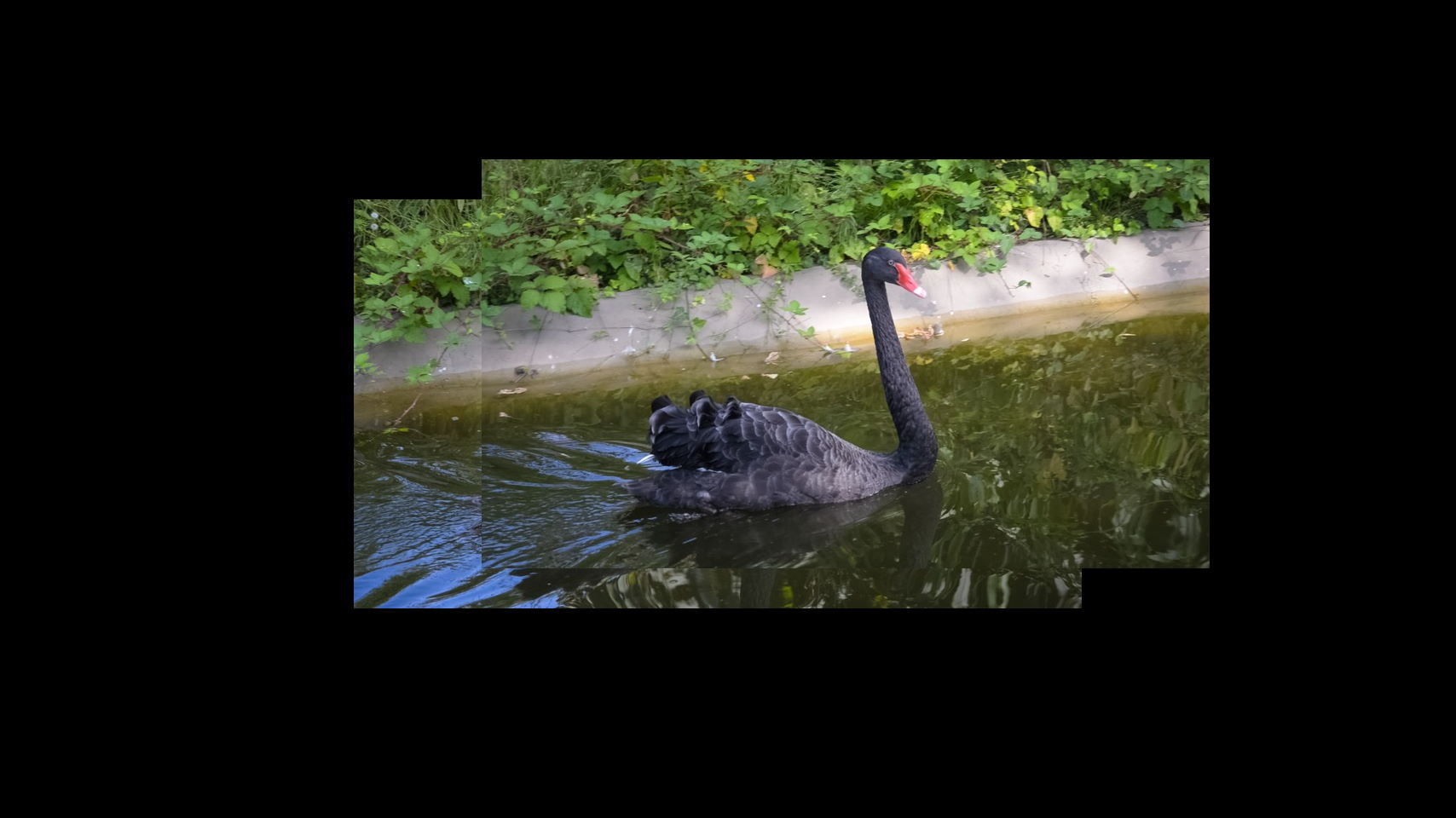}
\end{subfigure}%
\begin{subfigure}{.31\linewidth}
  \centering
  \includegraphics[width=.95\linewidth,valign=c]{figures/embedding/50_pct_overhead/embedded_input_frame2.jpg}
\end{subfigure}


\begin{subfigure}{.04\linewidth}
    b
\end{subfigure}%
\begin{subfigure}{.31\linewidth}
  \centering
  \includegraphics[width=.475\linewidth,valign=c]{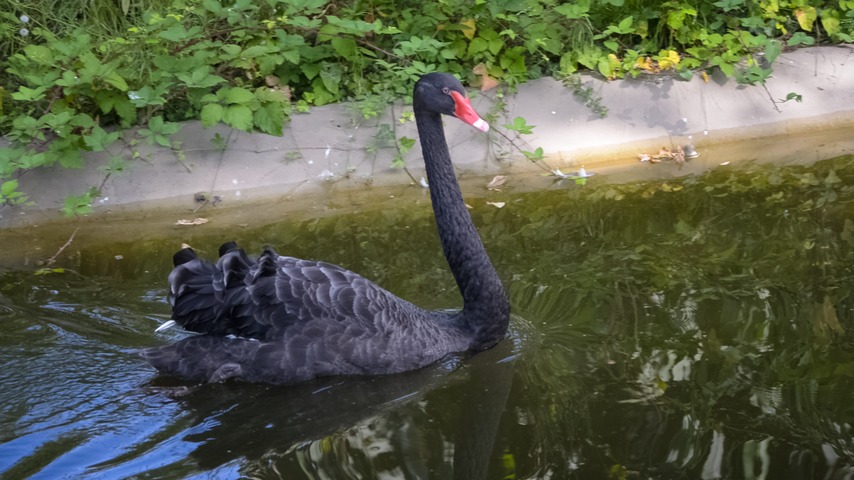}
\end{subfigure}%
\begin{subfigure}{.31\linewidth}
  \centering
  \includegraphics[width=.475\linewidth,valign=c]{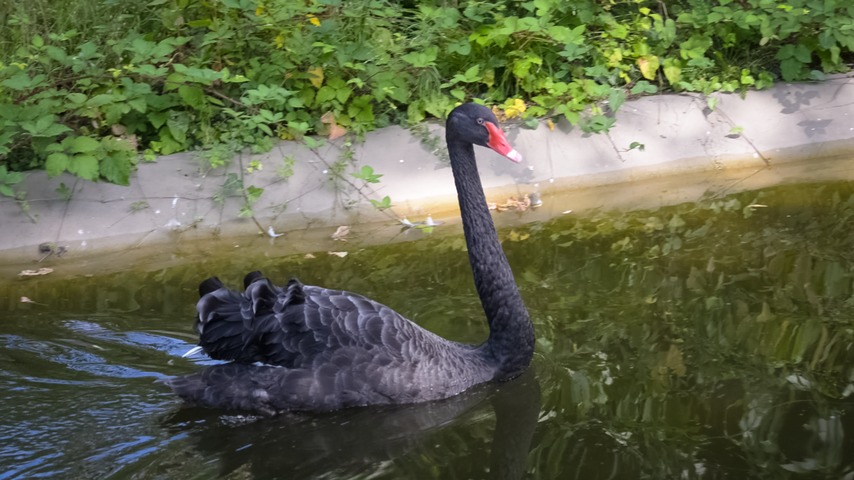}
\end{subfigure}%
\begin{subfigure}{.31\linewidth}
  \centering
  \includegraphics[width=.475\linewidth,valign=c]{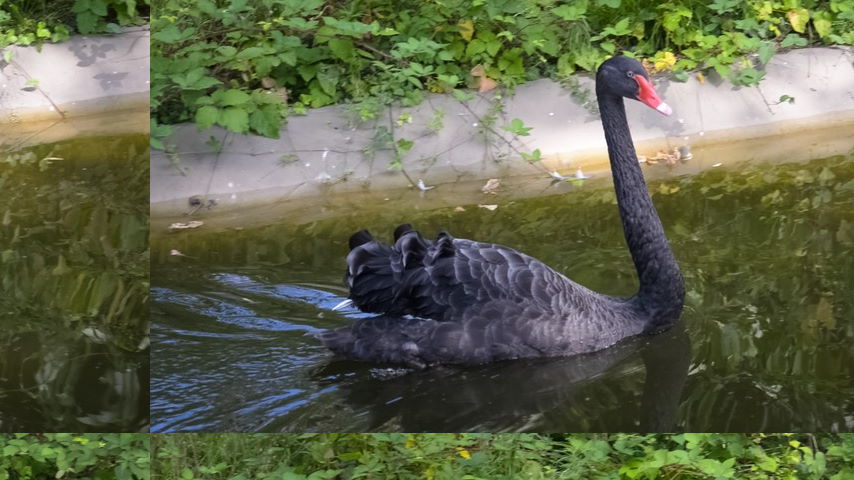}
\end{subfigure}%

\begin{subfigure}{.04\linewidth}
  \centering
  \caption*{}
\end{subfigure}%
\begin{subfigure}{.31\linewidth}
  \centering
  \caption*{Network Input 1}
\end{subfigure}%
\begin{subfigure}{.31\linewidth}
  \centering
  \caption*{Network Input 2}
\end{subfigure}%
\begin{subfigure}{.31\linewidth}
  \centering
  \caption*{Accumulated Input}
\end{subfigure}%

\caption{
To support moving camera input in sparse inference of frame differences,
a na\"ive approach (a) is to embed the input onto a larger image and only overwrite pixels that are covered in the next frame.
This approach comes with large overhead in memory consumption (storing the additional pixels of the embedded frame) and computational cost (checking all pixels for updates).
\ours (b) uses original shape buffers and feature maps -- avoiding the memory overhead and reducing the number of pixels that have to be checked for updates.
}
\label{fig:embedding_50_pct}
\end{figure*}

Sparse convolutions, on the other hand, can be used with existing pre-trained models, accelerating the large network to the speed of a smaller network without impacting prediction accuracy significantly \cite{Pan2018,Parger_2022}.
The linearity of convolutions enables the accumulation of updates over consecutive frames. 
After processing the first frame densely, upcoming frames can be processed using the difference between the current and the previous frame, the \emph{Delta}, as the network input.
Assuming a static camera scenario, this results in large image regions with no frame-to-frame difference.
Static elements like background or stationary objects do not require updates after the initial frame.
Researchers exploited this sparsity in previous work to reduce the number of FLOPs required for CNN inference \cite{Pan2018, Cavigelli2020, Habibian_2021_CVPR, Alwis2021, Dutson2021}, but were mostly unable to accelerate inference in practice.
With the recent progress in DeltaCNN \cite{Parger_2022}, theoretical FLOP reductions were translated into practical speedups on GPUs, using a custom sparse CNN implementation to exploit sparsity in all layers of the CNN.

DeltaCNN achieves speedups on multiple tasks and datasets, but like prior work, it is optimized for static camera inputs.
Even small frame-to-frame camera motion necessitates reprocessing large portions, or the entire image.
Spatially aligning consecutive frames, e.g., by leveraging camera extrinsics from IMUs or SLAM on mobile devices, can increase per-pixel similarity and thus update sparsity.
Due to the change of camera location and orientation, an aligned new frame may ``grow" beyond the initial field of view -- out of the previous-results buffers. 
One way to solve this issue is to embed inputs in a larger frame by padding the input image and drawing updates on top of it.
However, the downside of this approach is that padded input images increase the memory consumption and computational cost significantly, as shown in Figure \ref{fig:embedding_50_pct}. 

In this work, we propose \ours, a sparse CNN inference framework that allows moving camera input with marginal memory overhead.
Compared to previous work, we achieve up to 90\% higher frame rates in videos with moving cameras, 
indicating a new perspective for CNN inference optimization on low-power devices, such as surveillance cameras, smartphones, or VR headsets.

Our main contributions are:
\begin{itemize}
    \item We propose \ours, the first framework that leverages temporal continuity to accelerate CNN inference for videos with \emph{moving} cameras by processing only the sparse frame differences. 
    \item At the core of this work, we propose \emph{padded convolutions} for seamless integration of newly unveiled pixels without the need of reprocessing seen pixels.
    \item We design a two-dimensional ring buffer, a \emph{spherical buffer}, with wrapped coordinates. Our buffer allows partial growth, reset and initialization of new tiles without additional memory allocation. 
    \item We show how \ours can also be used for speeding up applications with static cameras when only parts of the image require processing. 
\end{itemize}
\begin{figure*}[pt]
\begin{subfigure}{.05\linewidth}
    F1
\end{subfigure}%
\begin{subfigure}{.190\linewidth}
  \centering
  \includegraphics[width=.95\linewidth,valign=c]{figures/embedding/50_pct_overhead/original_input_frame1.jpg}
\end{subfigure}%
\begin{subfigure}{.190\linewidth}
  \centering
  \includegraphics[width=.95\linewidth,valign=c]{figures/embedding/50_pct_overhead/original_input_frame1.jpg}
\end{subfigure}%
\begin{subfigure}{.190\linewidth}
  \centering
  \includegraphics[width=.95\linewidth,valign=c]{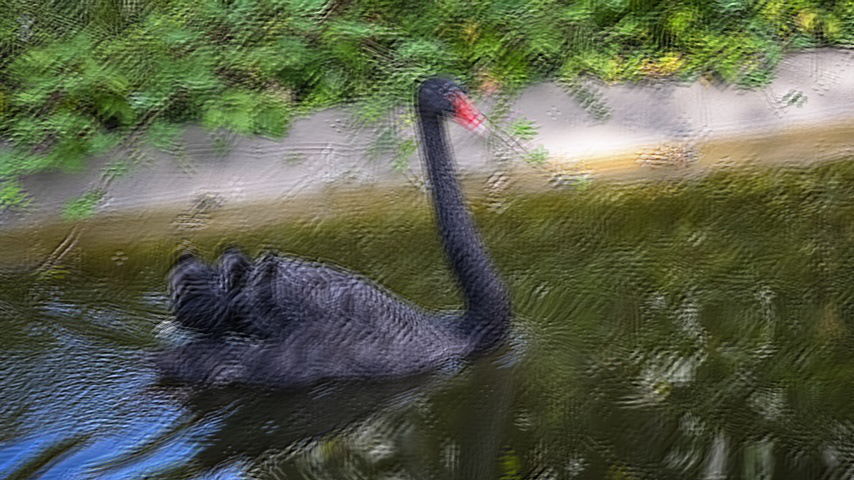}
\end{subfigure}%
\begin{subfigure}{.190\linewidth}
  \centering
  \includegraphics[width=.95\linewidth,valign=c]{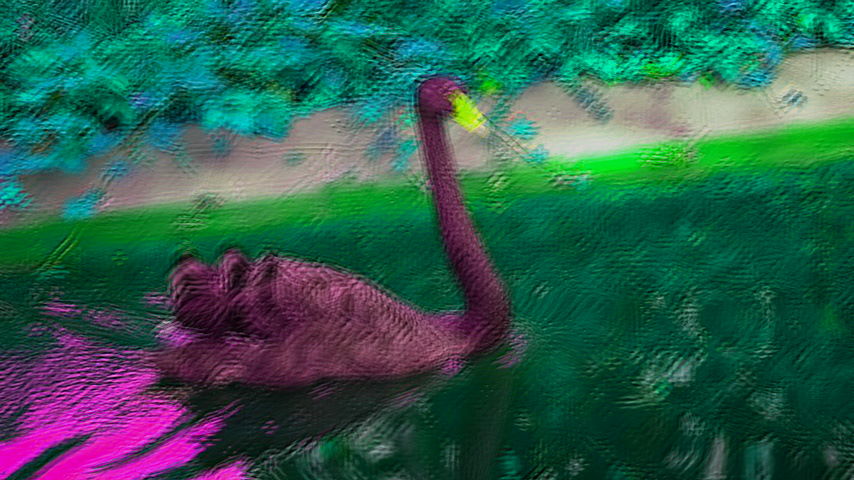}
\end{subfigure}%
\begin{subfigure}{.190\linewidth}
  \centering
  \includegraphics[width=.95\linewidth,valign=c]{figures/deltacnn_moving_cam_features/frame_1_conv_bias.jpg}
\end{subfigure}

\begin{subfigure}{.05\linewidth}
    F2
\end{subfigure}%
\begin{subfigure}{.190\linewidth}
  \centering
  \includegraphics[width=.95\linewidth,valign=c]{figures/embedding/50_pct_overhead/original_input_frame2.jpg}
\end{subfigure}%
\begin{subfigure}{.190\linewidth}
  \centering
  \includegraphics[width=.95\linewidth,valign=c]{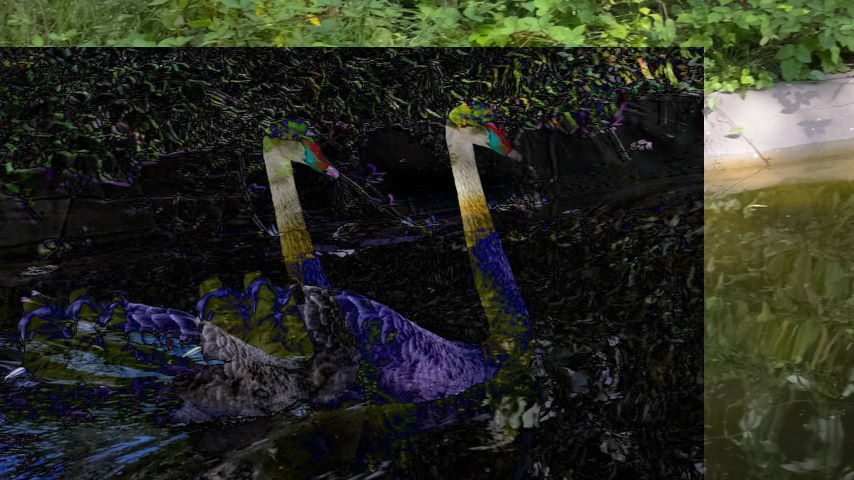}
\end{subfigure}%
\begin{subfigure}{.190\linewidth}
  \centering
  \includegraphics[width=.95\linewidth,valign=c]{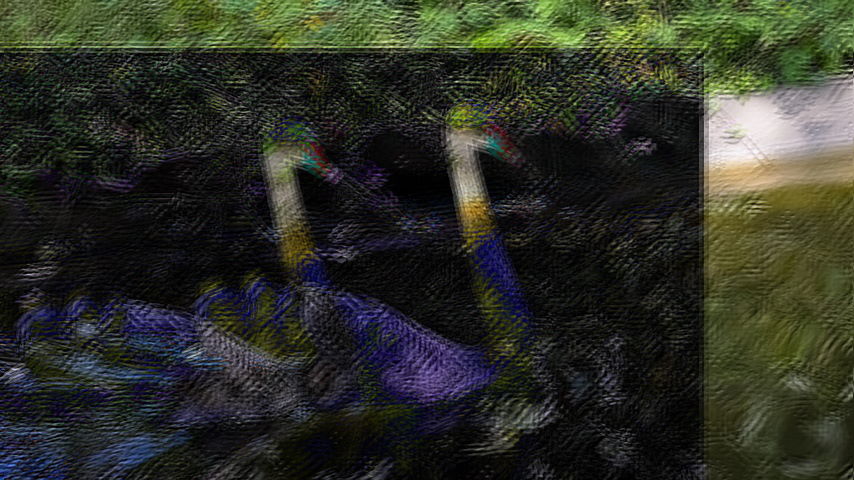}
\end{subfigure}%
\begin{subfigure}{.190\linewidth}
  \centering
  \includegraphics[width=.95\linewidth,valign=c]{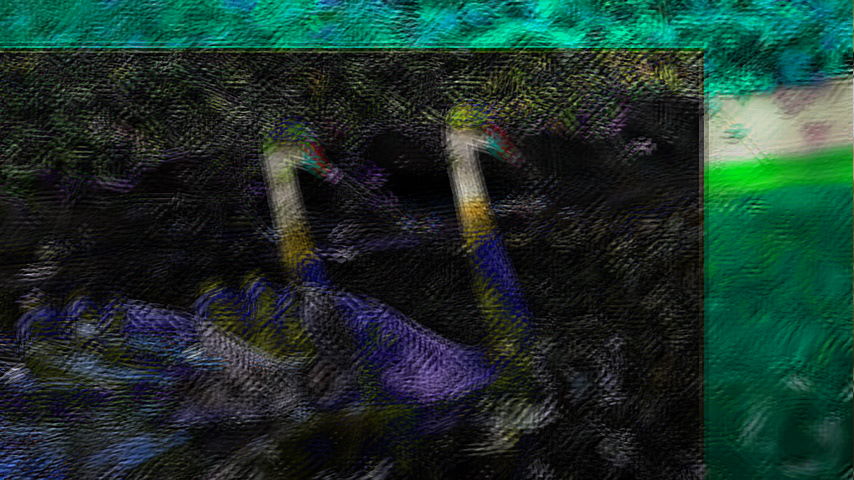}
\end{subfigure}%
\begin{subfigure}{.190\linewidth}
  \centering
  \includegraphics[width=.95\linewidth,valign=c]{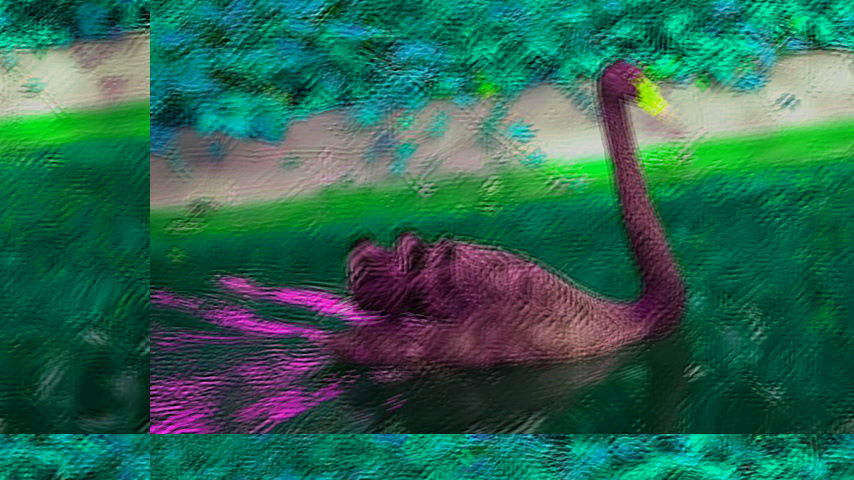}
\end{subfigure}

\begin{subfigure}{.05\linewidth}
  \centering
  \caption*{}
\end{subfigure}%
\begin{subfigure}{.190\linewidth}
  \centering
  \caption*{Input}
\end{subfigure}%
\begin{subfigure}{.190\linewidth}
  \centering
  \caption*{Delta}
\end{subfigure}%
\begin{subfigure}{.190\linewidth}
  \centering
  \caption*{Convolution}
\end{subfigure}%
\begin{subfigure}{.190\linewidth}
  \centering
  \caption*{+ Bias}
\end{subfigure}%
\begin{subfigure}{.190\linewidth}
  \centering
  \caption*{Accumulated Output}
\end{subfigure}

\caption{
The main concepts of \ours. We align frame 2 (F2) with frame 1 (F1) using homography matrices. 
For the intersecting region, we propagate only the aligned frame-to-frame differences (Delta). The newly uncovered regions are propagated directly.
The regions in which the buffers store the uncovered regions are reset to zero before processing the current Delta input.
After convolving frame 2, we add the bias to all previously unseen regions, and accumulate the result onto our spherical buffer using the offset coordinates for the current frame. 
}
\label{fig:deltacnn_features}
\end{figure*}

\section{Related Work}
The idea of only processing updated pixels in a convolutional layer has been proposed in various ways, with \emph{Recurrent Residual Module (RRM)} \cite{Pan2018} being the first work to apply this concept to videos.
RRM uses the difference between current and previous input, the \emph{Delta}, as input to convolutional layers. 
RRM demonstrates large theoretical reductions in FLOPs compared to dense inference resulting from skipping computations involving (nearly) zero-valued entries in feature maps.
In practice, skipping individual values is infeasible on most inference hardware like GPUs due to their single instruction, multiple data (SIMD) design.
Furthermore, RRM only processes convolutional layers on sparse Delta input.
All remaining layers are processed on the full, dense feature maps and therefore require expensive value-conversions before and after each convolutional layer.

\emph{Skip-Convolution} \cite{Habibian_2021_CVPR} and \emph{CBInfer} \cite{Cavigelli2020} improve the concept of RRM using a per-pixel sparse mask instead of per-value. 
Like RRM, CBInfer uses a threshold to truncate insignificant updates in the input feature map.
This threshold can be tuned for each layer for maximum sparsity while keeping the accuracy at a desired level.
In contrast, Skip-Convolution decides per \emph{output} pixel whether an update is required or can be skipped. 
In both cases, only the convolutional layers (and pooling layers in case of CBInfer) are processed sparsely, requiring expensive conversions between dense and sparse features before and after each of these layers, leaving large performance potential unused.

\emph{DeltaCNN} \cite{Parger_2022} propagates sparse Delta features end-to-end using a sparse Delta feature map together with an update mask.
This greatly reduces the memory bandwidth compared to previous work.
Propagating sparse updates end-to-end eliminates the necessity of converting the feature maps from dense accumulated values to sparse Deltas, and it speeds up other layers bottlenecked by memory bandwidth like activation, batch normalization, upsampling, etc.
Using a custom CUDA implementation, DeltaCNN outperforms cuDNN and CBInfer many times on video input.
However, DeltaCNN, as well as all other previous work, assumes static camera input.
Even single pixel camera motion between two frames can result in nearly dense updates when the image contains high frequency features.

\emph{Event Neural Networks} show that computational savings are strongly impacted by the intensity of camera motion\cite{Dutson2021}.
With slightly shaking cameras, theoretical FLOP savings were reduced by 40\% compared to static camera videos, while moving cameras reduced them by 60\%.
Depending on the distribution of the pixel updates, the impact on real hardware is likely even higher.

\emph{Incremental Sparse Convolution} uses sparse convolutions to incrementally update 3D segmentation masks online \cite{Liu2022}.
They solve a similar problem of reusing results from previously segmented regions, while allowing for new regions to be attached on the fly.
Due to their reliance on non-dilating 3D convolutions \cite{Graham2018}, attaching new regions leads to artifacts along the region borders.
We solve this issue using a standard, dilating convolution and by processing all outputs that are affected by the current input.

We propose \ours: building upon DeltaCNN, we design and implement a sparse inference extension that supports moving cameras.
Compared to DeltaCNN, we add the support for variable resolution inputs, spherical buffers, dynamic buffer allocation \& initialization and padded convolutions for seamless attachment of newly unveiled regions. 

\section{Method}
\ours relies on the following concepts:

    \textbf{Frame Alignment} The current frame is aligned with the initial frame of the sequence to maximize the overlap of consistent features (see Figure \ref{fig:deltacnn_features} \emph{Delta}).
    
    \textbf{Spherical Buffers} We use wrapped buffer offset coordinates in all non-linear layers to align them with the input (see Figure \ref{fig:deltacnn_features} \emph{Accumulated Output}).
    
    \textbf{Dynamic Initialization} When a new region is first unveiled due to camera motion, \ours adds biases onto the feature maps on the fly to all newly unveiled pixels to allow for seamless integration with previously processed pixels (see Figure \ref{fig:deltacnn_features} \emph{Bias}). 
    
    \textbf{Padded Convolutions} We add additional padding to convolutional layers to process all pixels that are affected by the kernel. These \emph{dilated} pixels are stored in truncated values buffers to enable seamless connection of potentially unveiled neighbor regions in upcoming frames (see Figure \ref{fig:dilated_padded_convolution}).

\subsection{Background: DeltaCNN}
For a better understanding of our contributions, we first summarize the core concept of DeltaCNN, 
which accelerates CNN inference by processing only the first frame densely and applying sparse updates to changed pixels in subsequent frames.
A sparse input {\em Delta} is calculated by subtracting the previously processed input from the current input frame.
Pixels whose Delta is smaller than a threshold are truncated to zero to increase sparsity.  

Because of linearity, the output of any convolution layer can be updated by convolving the sparse input Delta, $c(x + \delta) = c(x) + c(\delta)$.  This does not hold for non-linear layers like activation (\eg, ReLU) and pooling, which therefore require a separate \emph{accumulated values buffer} to accumulate the inputs to the respective layers across all previously processed frames. 
The buffers contain the full, dense layer input of the previous frame, including biases from previous layers.

Convolutions with kernel sizes larger than 1x1 pixels dilate the updates across the feature map.
A single pixel update in the center of the feature map might span over the entire feature map a few convolutional layers later.
This can lead to nearly dense updates for the larger part of the network.
DeltaCNN solves this by truncating insignificant updates not only on the input image, but at every activation layer.
To avoid accumulating errors indefinitely, the truncated values are accumulated in a dedicated \emph{truncated values buffer}.
Once an accumulated, truncated value exceeds a significance threshold, the value is propagated to the next layer. 

\subsection{Frame Alignment}
Robust frame alignment (or video stabilization) is a fundamental requirement for \ours to achieve high coherence across consecutive frames.
For best alignment between two frames in a 3D space, we use homography matrices with 8 degrees of freedom.
Homography matrices align two images in 3D to match a selected set of key features of the initial frame - typically the static background.

We embed the image in a grid-aligned frame to ensure a pixel-perfect downscale to all layers.
For example, in HRNet~\cite{Sun2019}, the highest resolution feature map spans 384x384 pixels, whereas the lowest resolution feature map spans only 12x12 pixels, resulting in a scaling factor of 32x32.
To ensure that all pixels align on all scaling levels, the input to HRNet must be aligned onto a grid of 32x32 pixel tiles. 
\Eg, if a frame has a 20-pixel horizontal offset relative to the first frame, we allocate 32 additional pixels, paste the original frame into the grid-aligned frame and set the 20 left-most pixels as well as the 12 right-most pixels to zero.

\begin{figure}[ptb]
\begin{subfigure}{.249\linewidth}
  \centering
  \includegraphics[width=.95\linewidth,valign=c]{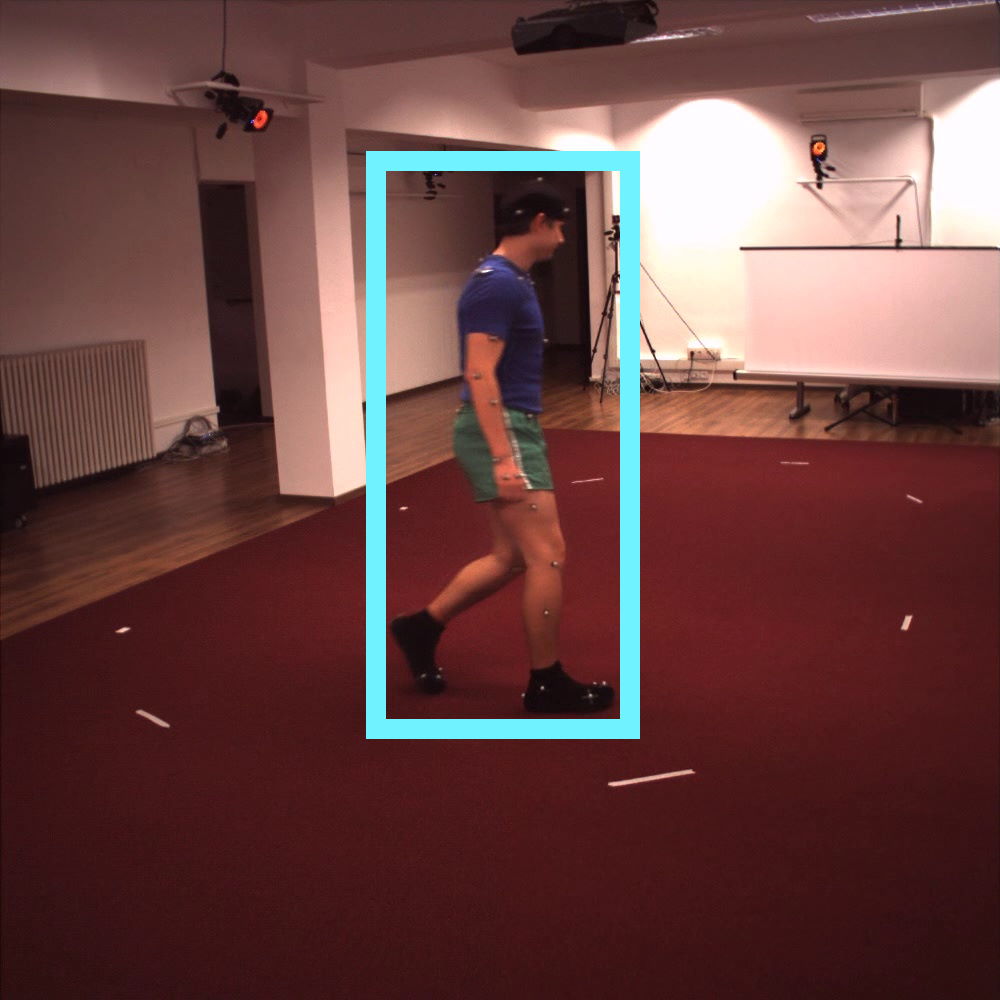}
  \caption*{Frame 1}
\end{subfigure}%
\begin{subfigure}{.249\linewidth}
  \centering
  \includegraphics[width=.95\linewidth,valign=c]{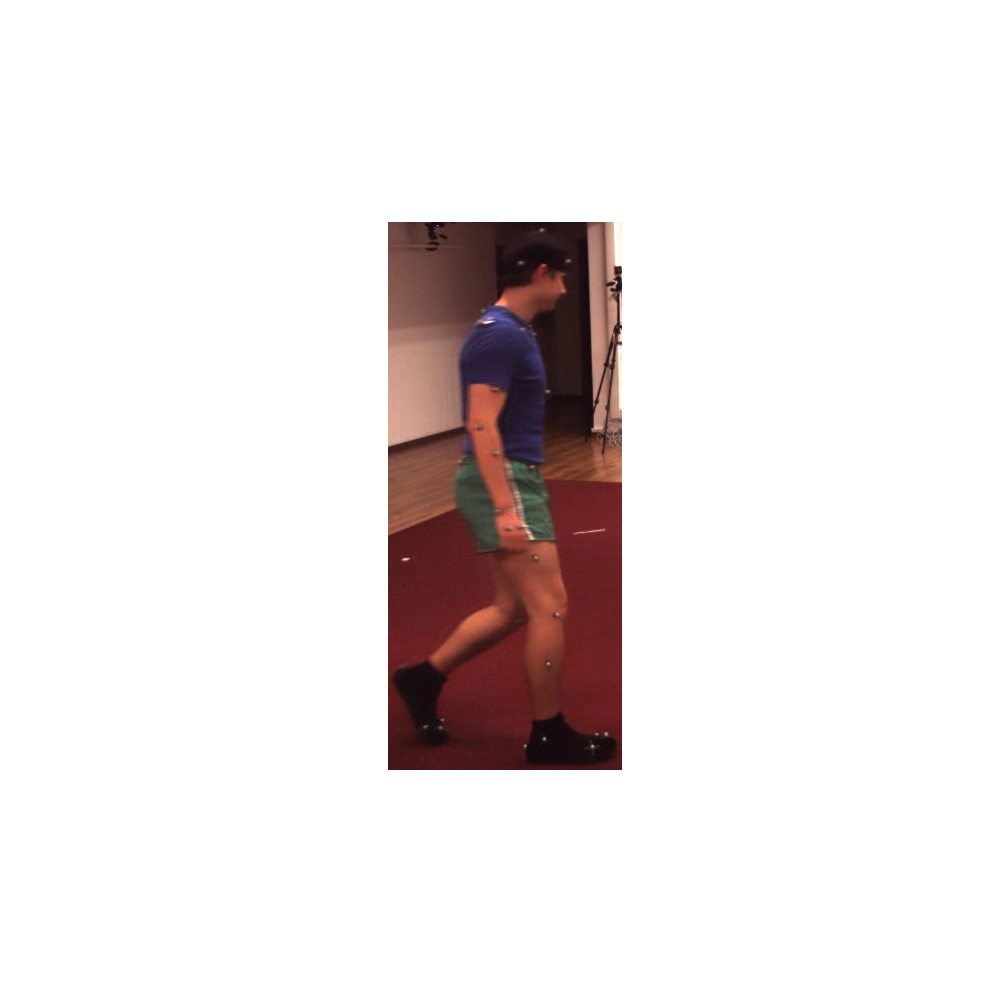}
  \caption*{Cropped}
\end{subfigure}%
\begin{subfigure}{.249\linewidth}
  \centering
  \includegraphics[width=.95\linewidth,valign=c]{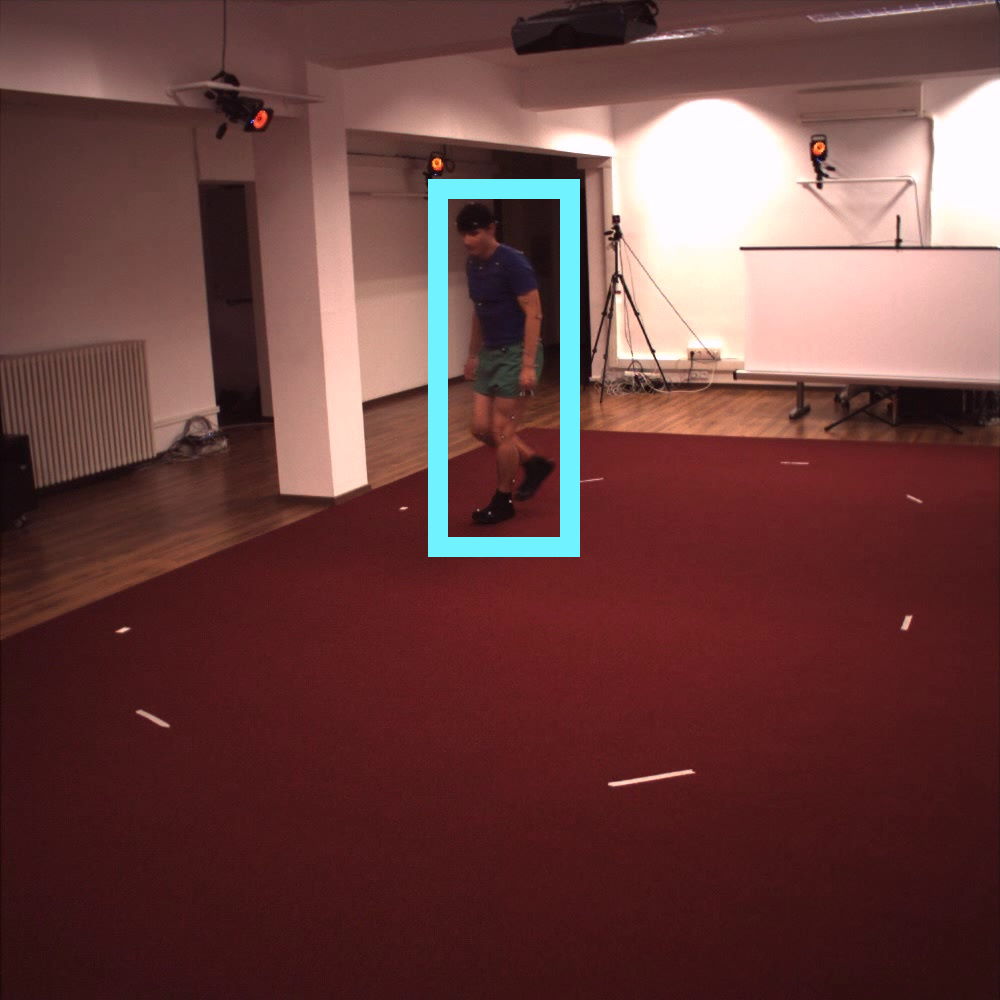}
  \caption*{Frame 2}
\end{subfigure}%
\begin{subfigure}{.249\linewidth}
  \centering
  \includegraphics[width=.95\linewidth,valign=c]{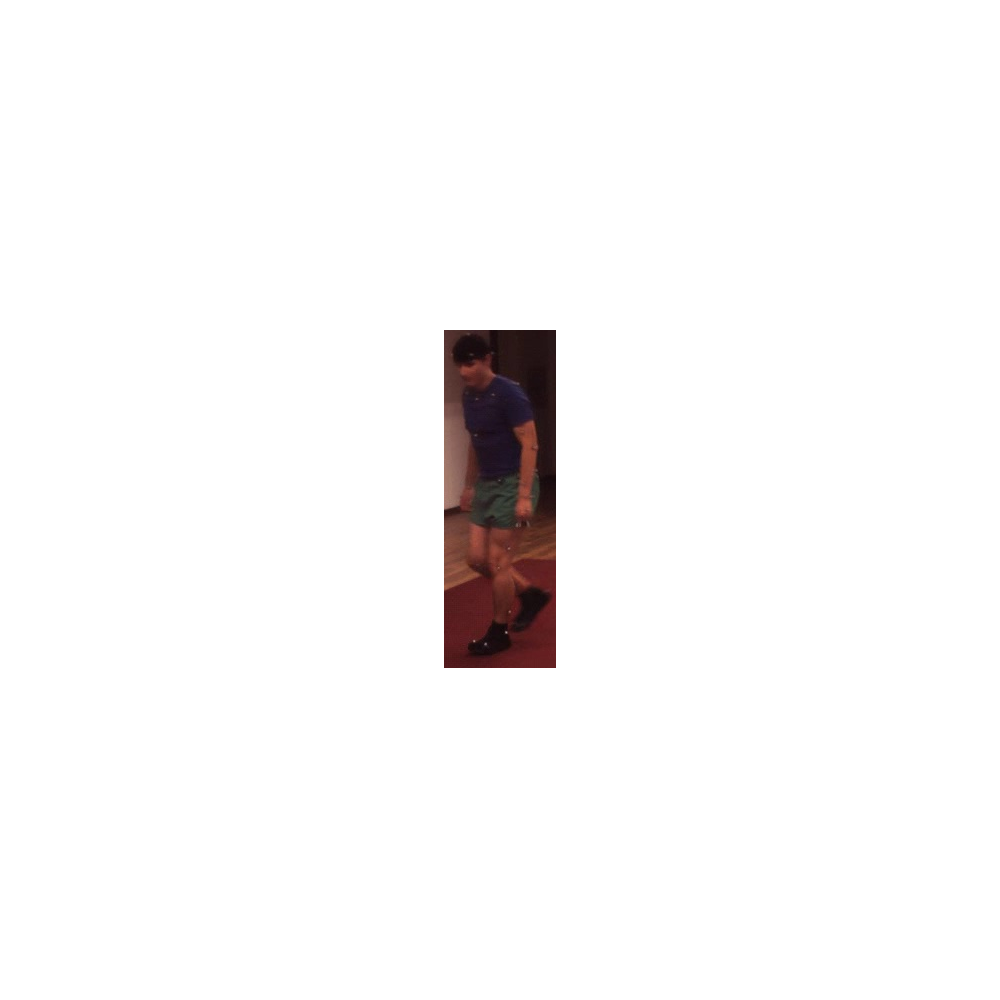}
  \caption*{Cropped}
\end{subfigure}%

\caption{
\ours can handle sparse Delta inputs of different crop sizes, and focus only on a fraction of the image, thanks to the concept of spherical buffers. This is not possible for DeltaCNN which always processes sparse Delta inputs of the entire frame. 
}
\label{fig:scaled_input}
\end{figure}
\subsection{Spherical Buffers}
\ours enables support for moving cameras by introducing 2D ring buffers, \emph{spherical buffers}.
These buffers align previously processed pixels with pixels of the current frame by translating feature map coordinates to buffer coordinates.
Pixel coordinates exceeding image borders are wrapped around the buffer, \ie, mapped to the opposite border using the modulo operator.
This enables the reuse of previous results for moving cameras without increasing the memory overhead over DeltaCNN (see Figure \ref{fig:dilated_padded_convolution}).
Our buffers support 2D offsets relative to the initial frame as well as cropped inputs (see Figure~\ref{fig:scaled_input}).
Frame scales can be helpful when only a part of the image is updated in the current frame - for example, when a frame is cropped to the bounding box of a region of interest. 
Tile offset indices are provided during inference to align features in buffers with the current input.
By multiplying tile offset indices with a layer's tile size (in the example above 32x32 pixels in highest resolution feature maps, 1x1 pixel in the lowest resolution), we get the pixel offset coordinates for buffer access in non-linear layers. 

\subsection{Dynamic Initialization}
When the input frame covers a previously unseen area, new tiles are allocated in all buffers before inferring the frame.
A new tile is allocated by deleting a previously stored tile at the desired location and resetting respective values to zero. 
In doing so, we lose information about the accumulated biases inferred in the initial frame.
We compensate for this by adding the biases onto reset regions of the Delta feature map during inference.

\subsection{Padded Convolutions}
\label{sec:padded_conv}
\begin{figure*}[pt]
\begin{subfigure}{.04\linewidth}
a
\end{subfigure}%
\begin{subfigure}{.155\linewidth}
  \centering
  \includegraphics[width=.95\linewidth,valign=c]{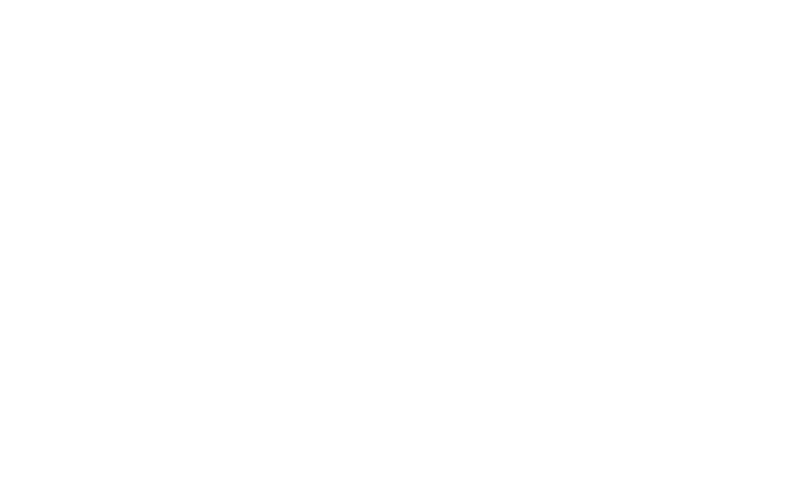}
\end{subfigure}%
\begin{subfigure}{.155\linewidth}
  \centering
  \includegraphics[width=.95\linewidth,valign=c]{figures/padding_2/blank.png}
\end{subfigure}%
\begin{subfigure}{.155\linewidth}
  \centering
  \includegraphics[width=.95\linewidth,valign=c]{figures/padding_2/blank.png}
\end{subfigure}%
\begin{subfigure}{.155\linewidth}
  \centering
  \includegraphics[width=.95\linewidth,valign=c]{figures/padding_2/blank.png}
\end{subfigure}%
\begin{subfigure}{.155\linewidth}
  \centering
  \includegraphics[width=.95\linewidth,valign=c]{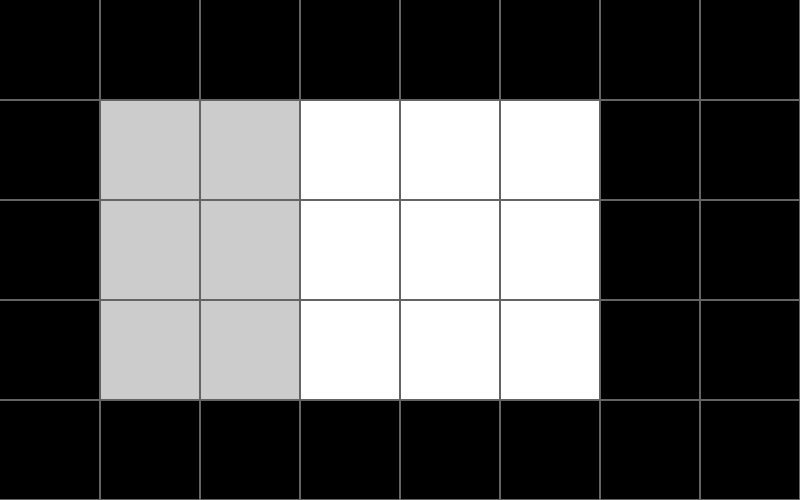}
\end{subfigure}%
\begin{subfigure}{.155\linewidth}
  \centering
  \includegraphics[width=.95\linewidth,valign=c]{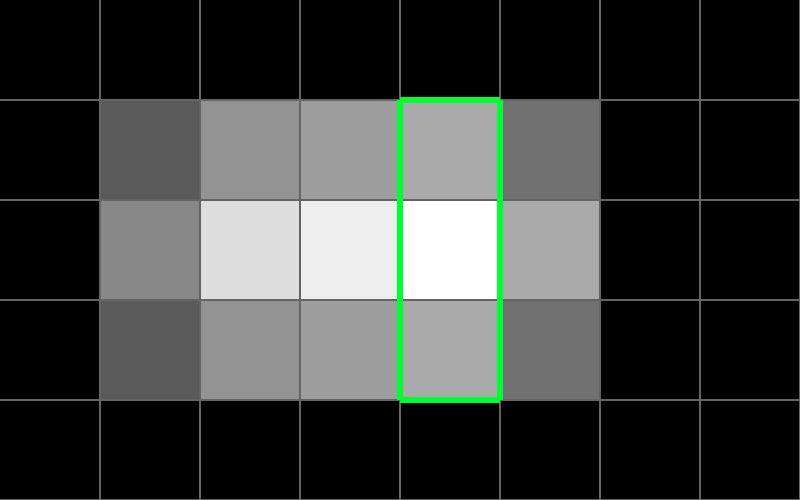}
\end{subfigure}

\begin{subfigure}{.04\linewidth}
b
\end{subfigure}%
\begin{subfigure}{.155\linewidth}
  \centering
  \includegraphics[width=.95\linewidth,valign=c]{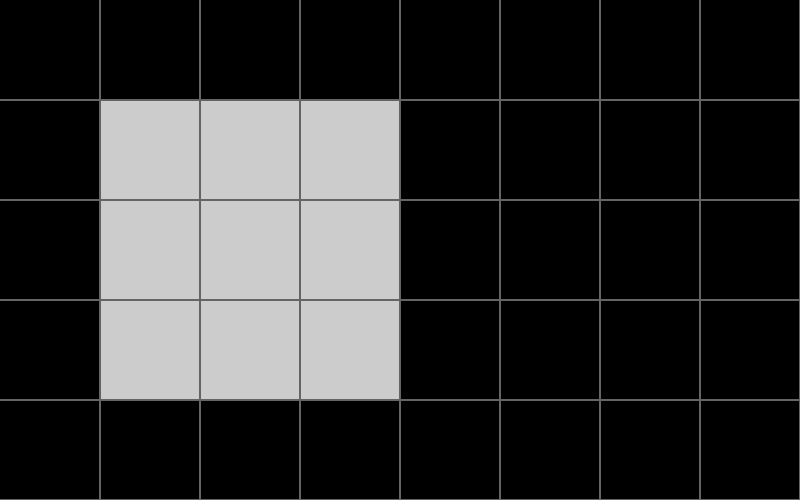}
\end{subfigure}%
\begin{subfigure}{.155\linewidth}
  \centering
  \includegraphics[width=.95\linewidth,valign=c]{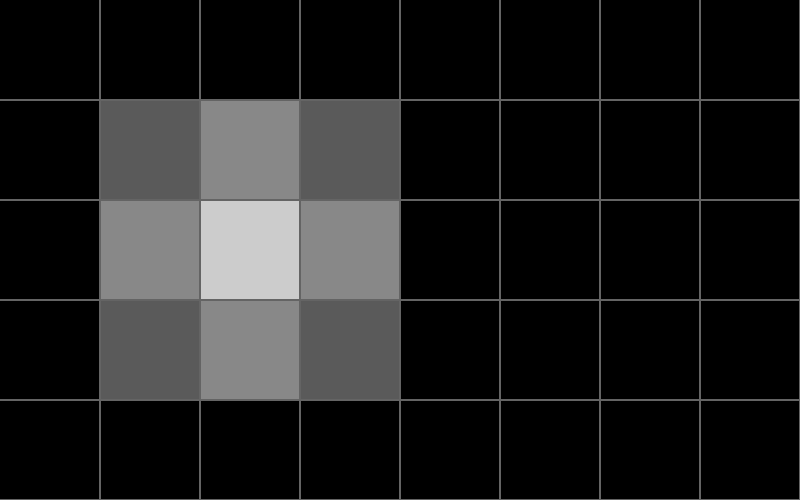}
\end{subfigure}%
\begin{subfigure}{.155\linewidth}
  \centering
  \includegraphics[width=.95\linewidth,valign=c]{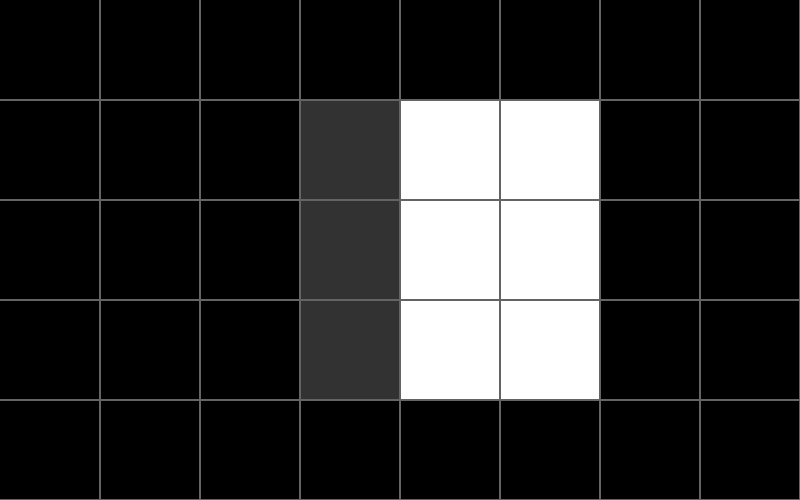}
\end{subfigure}%
\begin{subfigure}{.155\linewidth}
  \centering
  \includegraphics[width=.95\linewidth,valign=c]{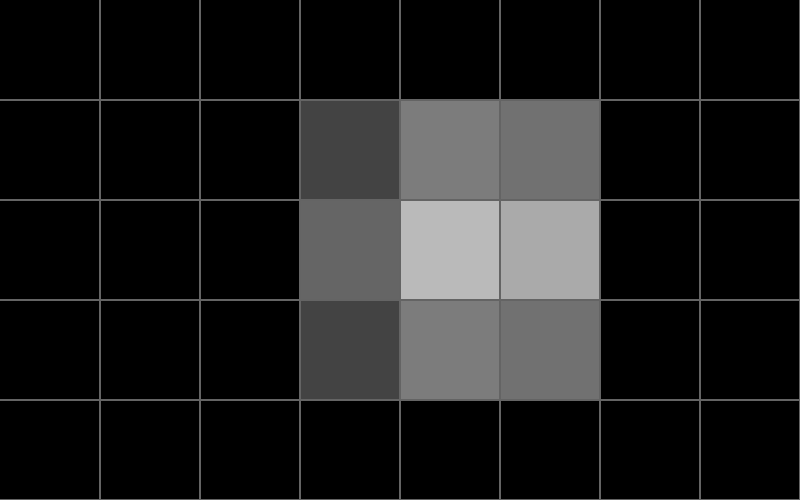}
\end{subfigure}%
\begin{subfigure}{.155\linewidth}
  \centering
  \includegraphics[width=.95\linewidth,valign=c]{figures/padding_2/frames_fused.png}
\end{subfigure}%
\begin{subfigure}{.155\linewidth}
  \centering
  \includegraphics[width=.95\linewidth,valign=c]{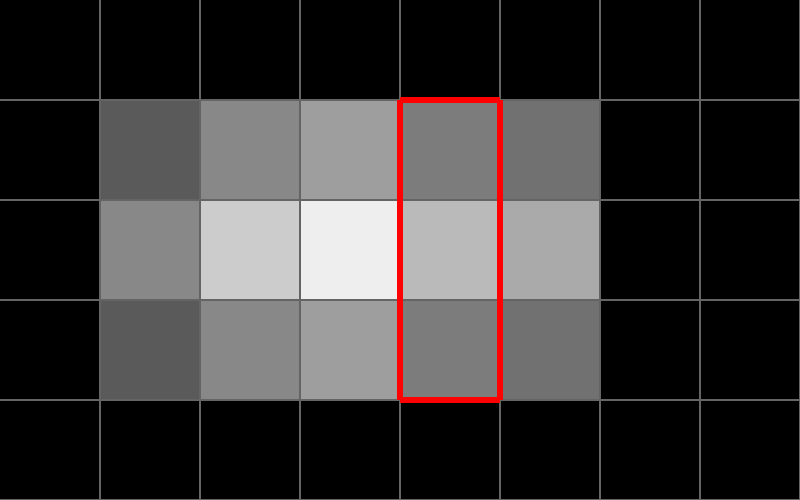}
\end{subfigure}%

\begin{subfigure}{.04\linewidth}
c
\end{subfigure}%
\begin{subfigure}{.155\linewidth}
  \centering
  \includegraphics[width=.95\linewidth,valign=c]{figures/padding_2/frame1.png}
\end{subfigure}%
\begin{subfigure}{.155\linewidth}
  \centering
  \includegraphics[width=.95\linewidth,valign=c]{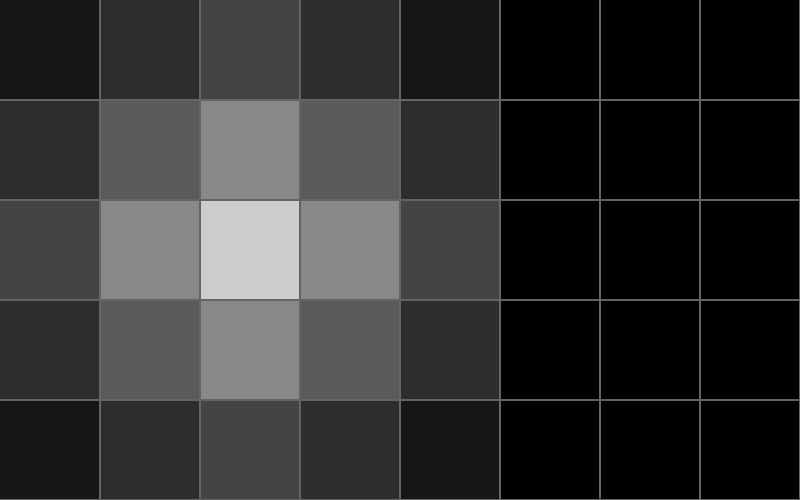}
\end{subfigure}%
\begin{subfigure}{.155\linewidth}
  \centering
  \includegraphics[width=.95\linewidth,valign=c]{figures/padding_2/delta2.png}
\end{subfigure}%
\begin{subfigure}{.155\linewidth}
  \centering
  \includegraphics[width=.95\linewidth,valign=c]{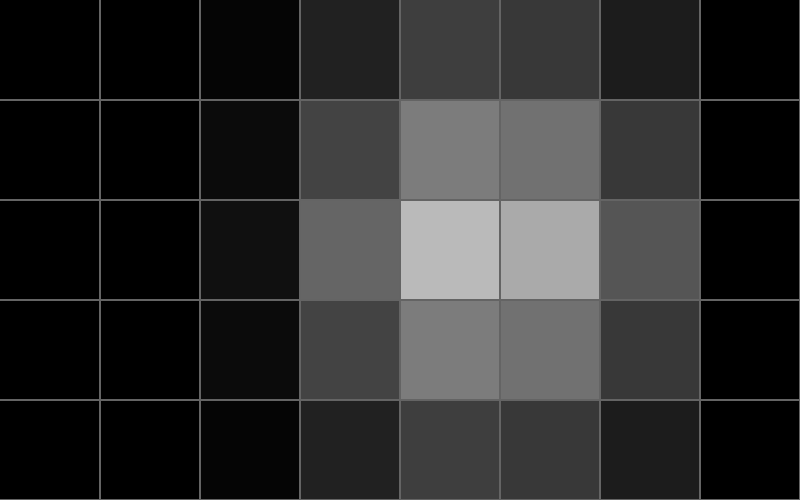}
\end{subfigure}%
\begin{subfigure}{.155\linewidth}
  \centering
  \includegraphics[width=.95\linewidth,valign=c]{figures/padding_2/frames_fused.png}
\end{subfigure}%
\begin{subfigure}{.155\linewidth}
  \centering
  \includegraphics[width=.95\linewidth,valign=c]{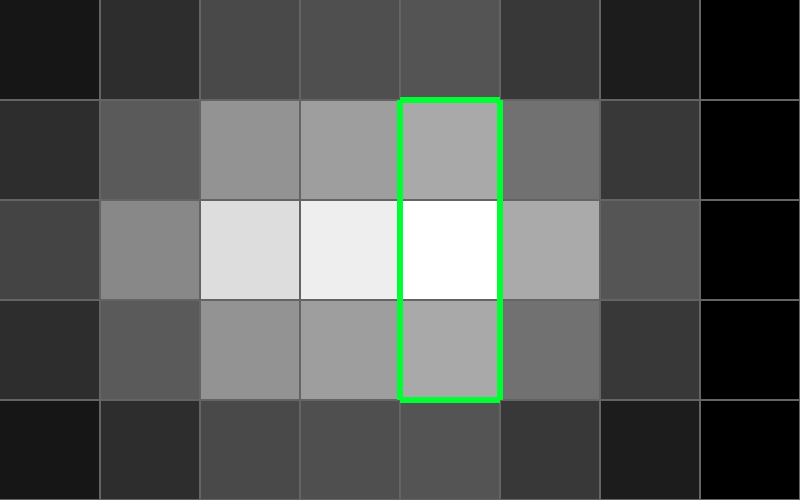}
\end{subfigure}%

\begin{subfigure}{.04\linewidth}
  \centering
  \caption*{}
\end{subfigure}%
\begin{subfigure}{.155\linewidth}
  \centering
  \caption*{Frame 1}
\end{subfigure}%
\begin{subfigure}{.155\linewidth}
  \centering
  \caption*{Output 1}
\end{subfigure}%
\begin{subfigure}{.155\linewidth}
  \centering
  \caption*{Delta Frame 2}
\end{subfigure}%
\begin{subfigure}{.155\linewidth}
  \centering
  \caption*{Output 2}
\end{subfigure}%
\begin{subfigure}{.155\linewidth}
  \centering
  \caption*{Accumulated Input}
\end{subfigure}%
\begin{subfigure}{.155\linewidth}
  \centering
  \caption*{Accumulated Output}
\end{subfigure}%

\caption{
Padded convolutions enable fusing images recorded at different perspectives.
Na\"ively applying convolutions separately to the initial frame and the Delta of Frame 2, and aggregating the outputs (row b) does not achieve the same results as one would get if the convolution is directly applied to a large fused input buffer (row a).
This is because of how the boundary pixels are treated and dropped when applying convolutions and border cropping. 
To address this issue, we need to cache and pad the output features, as shown in row c. 
See Section \ref{sec:padded_conv} for details regarding how to handle this in a practical manner.
}
\label{fig:dilated_padded_convolution}
\end{figure*}

After the initial frame, \ours propagates only Delta updates for the remaining sequence.
This can cause inconsistencies when mixing previously seen and newly unveiled pixels.
Consider a pair of \emph{adjacent} pixels $\mathbf{p}_{seen}$ and $\mathbf{p}_{new}$ across the boundary,
the resulting output at $\mathbf{p}_{new}$ may ``miss'' previously accumulated updates from pixel $\mathbf{p}_{seen}$~(see Figure~\ref{fig:dilated_padded_convolution}), since they were ``outside".

\ours addresses this {\em prospectively} by increasing the padding of convolutional layers to include all pixels that could be affected by convolutions on the current input.
This leads to an increased output size compared to the input when the kernel size is larger than 1x1.
For example, a 64x64 pixel input and a 3x3 kernel result in a 66x66 pixel output.

With modern CNNs consisting of hundreds of convolutional layers, growing the feature map at each layer could quickly exceed our buffers and computational budget.  To prevent this, in practice, we move and stash these \emph{output} values at \emph{dilated} pixels, and crop the output feature map of the current buffer to its original shape. 
Since \ours already uses truncated values buffers for storing values that were not yet propagated through the activation layer, we only need to increase the size of the buffer slightly to include the \emph{dilated} values and accumulate them there for later use.
Combined with spherical buffers, this allows us to fuse new image regions from all 4 directions seamlessly at the cost of memory and compute overhead for the \emph{dilated} border pixels.
Since most convolutions use kernels of size 3x3 and 1x1, the overhead is typically low.

\begin{figure*}[pt]
\begin{subfigure}{.165\linewidth}
  \centering
  \includegraphics[width=.95\linewidth]{figures/noise_suppression/00000.jpg}
  \caption*{Frame 1}
\end{subfigure}%
\begin{subfigure}{.165\linewidth}
  \centering
  \includegraphics[width=.95\linewidth]{figures/noise_suppression/00001.jpg}
  \caption*{Frame 2}
\end{subfigure}%
\begin{subfigure}{.165\linewidth}
  \centering
  \includegraphics[width=.95\linewidth]{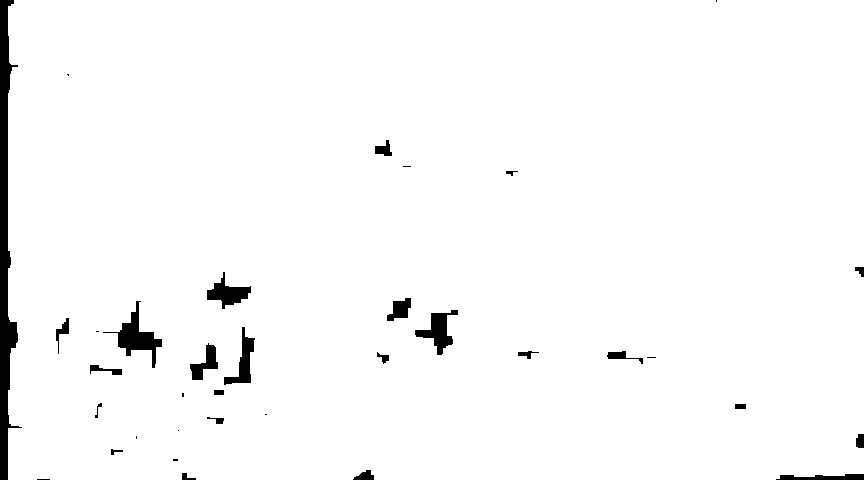}
  \caption*{T}
\end{subfigure}%
\begin{subfigure}{.165\linewidth}
  \centering
  \includegraphics[width=.95\linewidth]{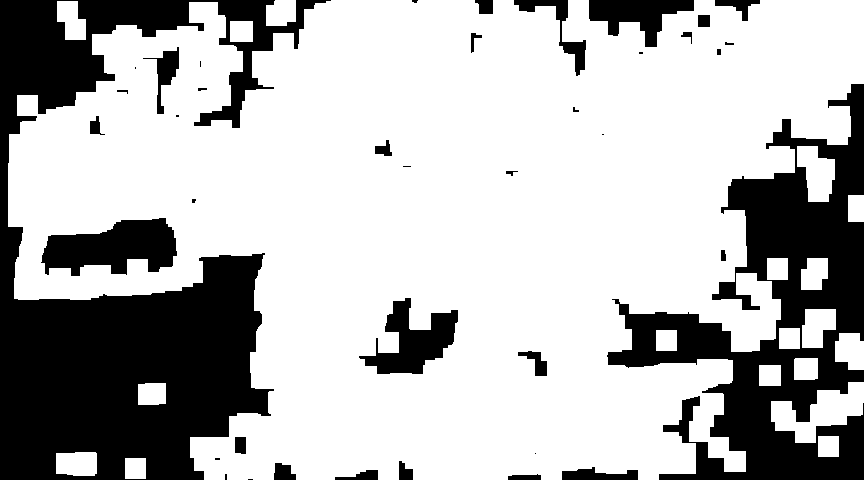}
  \caption*{T+S}
\end{subfigure}%
\begin{subfigure}{.165\linewidth}
  \centering
  \includegraphics[width=.95\linewidth]{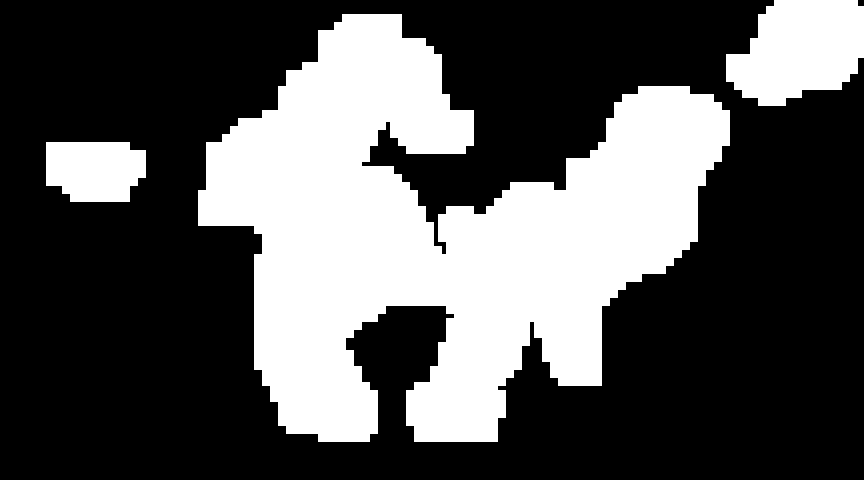}
  \caption*{T+N}
\end{subfigure}%
\begin{subfigure}{.165\linewidth}
  \centering
  \includegraphics[width=.95\linewidth]{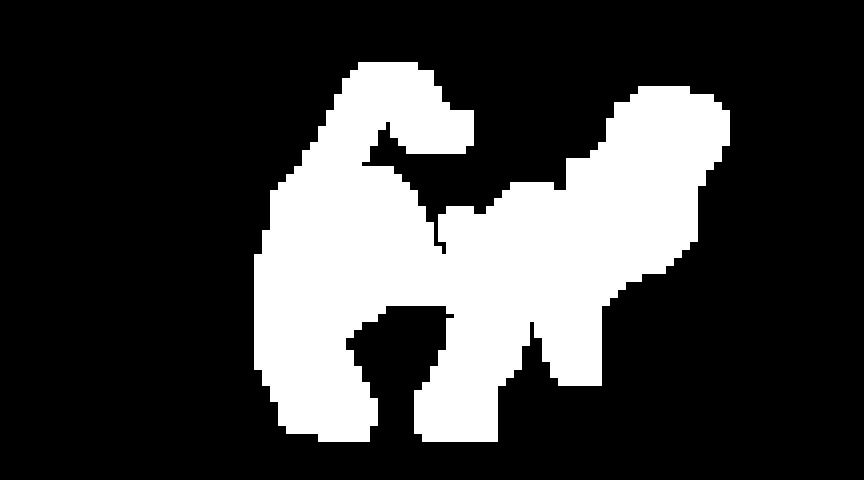}
  \caption*{T+S+N}
\end{subfigure}

\caption{
We increase update sparsity using noise suppression and region of interest scaling.
$T$ uses the original truncation implementation of DeltaCNN.
Scaling the Delta further away from the region of interest ($T+S$) slightly increases sparsity.
Using noise suppression ($T+N$) to eliminate single pixel updates increases sparsity even further.
Combining scaling and noise suppression ($T+S+N$) achieves the highest sparsity, capturing only the most important updates of the moving camel.
Note that in all cases, we dilate the update mask by 10 pixels; explaining the low sparsity in case $T$. Source: DAVIS 2017
}
\label{fig:noise_suppression}
\end{figure*}

\subsection{Buffer management}
We use a buffer manager that is responsible for: keeping track of each tile's state, allocating new regions in the spherical buffer, initializing newly allocated tiles and adding biases to feature maps during inference.
As mentioned before, spherical buffers wrap coordinates around image borders to allow for infinite panning. 
One downside of this approach is that when borders grow in one direction, we overwrite the \emph{dilated} pixels stored for growing in the opposite direction (see Figure \ref{fig:teaser}). 
Restoring \emph{dilated} pixels is non-trivial and would require inferring the negative accumulated values of the deleted tiles - which adds high overhead.
Instead, we simply delete the \emph{dilated} pixels and restrict the growth in the opposite direction. 
Once the camera moves back over a restricted region, all buffers are reset entirely, and the current frame has to be inferred densely - becoming the new reference until the next buffer reset.
Restoring the truncated values by only resetting the tiles next to the invalid truncated values does not work, as they would again lack valid truncated values required for seamless attachment to persistent tiles.
While a full reset reduces the reuse of previously processed features, in our evaluations, the number of resets is fairly small as camera pans typically last many frames without changing directions.
Our buffer manager stores for each tile of the grid the currently held global coordinates and the directions in which the growth is restricted.
Every time the camera moves, the buffer manager checks whether the tiles already hold a value from the corresponding coordinates. 
If not, it resets the respective tiles if possible or requests a buffer reset if growth is restricted. 

\subsection{Optimizations}
While the above-mentioned features can already lead to a large speedup in moving camera scenarios, the speedup can be increased significantly with some small optimizations.
For the DAVIS 2017 dataset \cite{Pont-Tuset_arXiv_2017}, we noticed strong salt-and-pepper style noise in the update masks.
Since video stabilization is not perfect, some pixels align incorrectly, especially in high frequency details like leaves of a tree. 
This leads to a strong color change for some isolated pixels in the image, resulting in a large compute overhead when processing the image in tiles.
We mitigate this issue using region-of-interest (ROI) focused sensing and noise suppression.
When we know the ROI from previous frames, \eg, in the case of object detection or object segmentation, we can increase the sensitivity around these regions and decrease the sensitivity further away.
Additionally, we suppress noise coming from imperfect image alignment by downscaling the Delta, processing which pixels exceed the thresholds and suppressing single pixel updates using average pooling.
Overall, these steps lead to a significant speedup with only minor impact on accuracy (see Figure \ref{fig:noise_suppression}).
Further optimizations can be found in the supplemental material.
\section{Evaluation}
\begin{figure*}[ptb]
    \centering
    \includegraphics[width=0.99\textwidth]{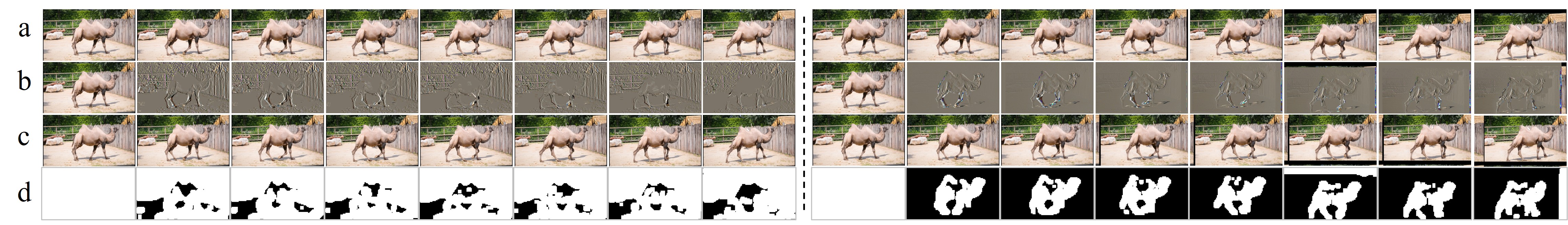}
    \caption{A DAVIS 2017 sequence processed using DeltaCNN (left) and \ours (right). Row a shows the original input on the left, and the aligned and cropped input on the right. Row b visualizes the Delta between the (aligned) input and the previous input in row c. Row d visualizes the update mask with white pixels indicating an update. \ours achieves much higher sparsity on the same input by reusing previous results for the aligned, static background.
    }
    \label{fig:results}
\end{figure*}
We evaluate \ours in two scenarios: a) using video stabilization to align subsequent frames with moving cameras and b) using bounding boxes in a static camera scenario to infer only regions of interest while retaining out-of-interest features from previous frames.

\textbf{Video Object Segmentation}
In task a), we evaluate the video object segmentation model BMVOS~\cite{Cho2022} on the DAVIS 2017 dataset \cite{Pont-Tuset_arXiv_2017}.
BMVOS uses the DenseNet~\cite{Huang2017} backend for feature encoding, the most computationally expensive part of the model.
We use weights pretrained on the DAVIS dataset, which were provided by the authors of BMVOS.
The DAVIS 2017 dataset consists of a set of sequences with one or more segmented objects of which the masks are provided only for the first frame - and the model needs to paint the segmentation mask for the remaining frames in the video.
Most of the videos are recorded using moving cameras, and are therefore well suited for evaluating \ours.
The DAVIS 2017 dataset does not provide camera extrinsics.
Instead, we use OpenCV~\cite{opencv_library} to estimate homography matrices between two consecutive frames.
This simplistic method fails in some cases, \eg, when it does not find good features due to motion blur or with motion parallax.
We see the tasks of video stabilization as orthogonal and therefore only evaluate on the set of videos (14 sequences) for which this simple video stabilization approach works reliably.

In video object segmentation, the segmentation mask from the previous frame can be used to suppress noise in the upcoming frame.
We use this mask to lower the update sensitivity further away from the ROI - leading to an increase in sparsity. 
We deliberately tuned the thresholds for DeltaCNN with a lower accuracy target than for \ours to show that we achieve higher frame rates even when targeting higher accuracy.
The details about threshold tuning and ROI-focused sensing are found in the supplemental material.


\textbf{Human Pose Estimation}
In task b), we perform human pose estimation using \emph{HRNet}~\cite{sun2019deep,xiao2018simple} on the \emph{Human3.6M}\footnote{The Human3.6M dataset was accessed and used only by the co-authors at Graz University of Technology} dataset~\cite{IonescuSminchisescu11,h36m_pami}.
Human3.6M uses static cameras and is recorded in a static studio with only a single moving actor in the room.
We use this dataset to evaluate a second scenario for \ours: top-down pose estimation. 
With top-down pose estimation, humans in the scene are detected first, and then the pose is predicted for each detected human.
Each bounding box is cropped and inferred individually - estimating one set of joints per person.
Since Human3.6M provides per frame bounding boxes of the actors as part of their labels, we omit the step of detecting humans in the scene.
We snap the given bounding boxes onto a 32x32 pixel grid and then increase them by 32 pixels on each side where possible.
The increased bounding boxes ensure that joint positions are not too close to image borders, as this could negatively impact prediction accuracy. 
We use the same weights and thresholds as were used in the DeltaCNN paper and evaluate on the same data subset (Subject S11).

Both scenarios are compared against dense inference of cuDNN and against the original DeltaCNN without moving camera support.
In HRNet, we only measure model inference time, without the overhead of post-processing or accuracy evaluation.
The input shape is 384x384 with a tile size of 32x32 pixels unless stated otherwise.
In the case of BMVOS, we only measure the duration of the feature encoding model to better highlight the difference of pure CNN inference speeds.
For this evaluation, the camera input is embedded into a frame of shape Wx480 pixels, where W depends on the aspect ratio of the video (typically 864px) and is aligned onto a grid with a tile size of 48x48 pixels.

\textbf{Hardware}
We conduct our evaluations on two GPUs: Nvidia GTX 1080 Ti and Nvidia RTX 3090 with 3584/10496 CUDA cores and 11GB/24GB of VRAM respectively. 
The evaluations are performed using 32-bit precision, but our approach can be applied to other data types as well.

\section{Results}
\begin{table*}[ptb]
\centering
\setlength{\tabcolsep}{4pt}
\renewcommand\arraystretch{1.2}
\resizebox{\linewidth}{!}{
\begin{tabular}{cc||c|c|cc|cc|cc|cc}
\hline\thickhline
\multirow{2}{*}{Videos} & \multirow{2}{*}{Backend} & \multirow{2}{*}{IoU} & \multirow{2}{*}{GFLOPs} & \multicolumn{2}{c|}{GTX 1080 Ti b=1} & \multicolumn{2}{c|}{GTX 1080 Ti b=5} & \multicolumn{2}{c|}{RTX 3090 b=1} & \multicolumn{2}{c}{RTX 3090 b=12} \\
\cline{5-12}
& & & & FPS & speedup & FPS & speedup & FPS & speedup & FPS & speedup \\
\hline
\multirow{5}{*}{All} 
& cuDNN                         & 76.85 & 22.1  & 30.7  & 1.0   & 40.9  & 1.0   & 45.4 & 1.0    & 103   & 1.0 \\
& DeltaCNN (dense)              & 76.85 & 22.1  & 35.0  & 1.14  & 41.2  & 1.01  & 66.5 & 1.46   & 100   & 0.97 \\
& \ours (dense)                 & 76.76 & 22.6  & 33.5  & 1.09  & 40.3  & 0.99  & 57.7 & 1.27   & 93.6  & 0.91 \\
& DeltaCNN (sparse)             & 76.18  & 15.5 & 42.5  & 1.38  & 51.4  & 1.26  & \textbf{70.0} & \textbf{1.54} & 125   & 1.21 \\
& \textbf{\ours (sparse)}       & 76.32 & \textbf{6.8} & \textbf{54.8}  & \textbf{1.79}  & \textbf{81.9}  & \textbf{2.00}  & 65.2 & 1.44    & \textbf{215} & \textbf{2.1} \\
\hline

\multirow{3}{*}{Moving} 
& cuDNN                         & 79.59 & 22.1  & 30.7  & 1.0   & 40.9  & 1.0   & 45.4 & 1.0   & 103   & 1.0 \\
& DeltaCNN (sparse)             & 79.38 & 18.8  & 38.2  & 1.24  & 44.1  & 1.08  & \textbf{70.2} & \textbf{1.55}   & 106     & 1.03 \\
& \textbf{\ours (sparse)}       & 79.43 & \textbf{7.3}  & \textbf{52.3}  & \textbf{1.70}  & \textbf{75.1} & \textbf{1.84}  & 65.0 & 1.43    & \textbf{201}     & \textbf{1.95} \\
\hline\thickhline
\end{tabular}
}
\caption{
Results for the task of video object segmentation using BMVOS on the DAVIS dataset. We track only the DenseNet backend for evaluating frames per second (FPS). For accuracy benchmarks, the remaining network is processed densely.
GFLOPs are reported as average over the test set. Both devices are evaluated with a batch size $b$ of one, and the maximum batch size each device can hold in memory.
}
\label{tab:acc_speedup_vos}
\end{table*}




We evaluate speedup and accuracy of \ours compared to DeltaCNN without explicit moving camera support as well as the dense cuDNN backend.
In this section, we report the differences on the GTX 1080 Ti - for results on the RTX 3090, please see the respective tables.

\textbf{Video Object Segmentation}
\ours outperforms state-of-the-art backends in sparse mode by a margin of up to 100\% over the cuDNN baseline  (see Table \ref{tab:acc_speedup_vos}).
The DAVIS 2017 dataset consists mainly of videos with large bounding boxes and often fast camera motion, leading to a high average update rate of 29\%.

Excluding videos without camera motion highlights the relative speedup over DeltaCNN without moving camera support with +85\% speedup over cuDNN instead of +8\%.

\ours, like DeltaCNN, requires larger workloads with more powerful hardware to show speedups.
With 10.496 cores in a RTX 3090, small batch sizes typically do not fully utilize the available hardware resources in most layers - skipping some tiles in this scenario does not affect the overall performance unless all tiles can be skipped altogether.
By increasing the batch size, we can fully utilize the GPU - cores that skip a tile can immediately continue processing the next tile, leading to an overall speedup.

Over the 990 frames in the DAVIS dataset, we had to reset the buffers 9 times because of cameras panning over invalid truncated values.
This low number of resets has only a small impact on the overall performance.

\begin{table*}[ptb]
\centering
\setlength{\tabcolsep}{4pt}
\renewcommand\arraystretch{1.2}
\resizebox{\linewidth}{!}{
\begin{tabular}{c||c|c|c|cc|cc|cc|cc}
\hline\thickhline
\multirow{2}{*}{Backend} & \multirow{2}{*}{PCKh@0.5} & \multirow{2}{*}{PCKh@0.2} & \multirow{2}{*}{GFLOPs}& \multicolumn{2}{c|}{GTX 1080 Ti b=1} & \multicolumn{2}{c|}{GTX 1080 Ti b=16} & \multicolumn{2}{c|}{RTX 3090 b=1} & \multicolumn{2}{c}{RTX 3090 b=40} \\
\cline{5-12}
& & & & FPS & speedup & FPS & speedup & FPS & speedup & FPS & speedup \\
\hline
cuDNN         & \multirow{2}{*}{96.03\%}   & \multirow{2}{*}{83.33\%} & \multirow{2}{*}{47.1}
                                                            & 22.5  & 1.0   & 24.8  & 1.0   & 12.3  & 1.0   & 69.6  & 1.0   \\
DeltaCNN (dense)              &           &           &       & 24.2  & 1.08  & 41.3  & 1.7   & 26.0  & 2.1   & 95.8  & 1.4   \\
\ours (dense)                  & 96.00\%   & 83.32\%   & 17.7     & 32.5  & 1.44  & 91.4  & 3.7   & 22.0  & 1.8   & 232   & 3.3   \\
cuDNN (cropped)               & 95.92\%   & 83.11\%   & 13.3  & 24.0  & 1.07  & 71.0  & 2.9   & 11.8  & 1.0   & 209   & 3.0   \\
DeltaCNN (sparse)             & 95.95\%   & 82.31\%   & 3.55  & \textbf{44.9}  & \textbf{2.0}   & 202   & 8.1   & \textbf{26.2}  & \textbf{2.1}   & 547   & 7.9   \\
\textbf{\ours (sparse)}       & 95.95\%   & 82.33\%   & 3.45  & 41.2  & 1.83  & 228   & 9.2   & 22.6  & 1.8   & 584   & 8.4   \\
\textbf{\ours (sparse \textdagger)}    & 95.93\%   & 82.22\%   & \textbf{3.44}  & 42.5  & 1.89  & \textbf{276}   & \textbf{11.1}  & 22.5  & 1.8   & \textbf{691}   & \textbf{9.9}   \\
\hline\thickhline
\end{tabular}
}
\caption{Speed and accuracy comparisons of different CNN backends used for pose estimation on the Human3.6M dataset. The same set of auto-tuned thresholds for update truncation is used for all devices and batch sizes $b$.
\textdagger~limits the size of the input from 384x384 to 288x288 pixels - freeing up 44\% of the memory used for buffers.
The freed-up memory is used to increase the batch size to fully utilize the available VRAM  (b=28 on 1080 Ti and b=66 on 3090).
cuDNN cropped uses the same, cropped input as ours, instead of the full camera input.
}
\label{tab:acc_speedup_pose_estimation}
\end{table*}

\textbf{Human Pose Estimation}
\ours can outperform cuDNN by up to 11x in the case of HRNet (see Table~\ref{tab:acc_speedup_pose_estimation}).
Please note that the frame rate of cuDNN is 30\% lower than in the DeltaCNN paper since we enforced the use of 32-bit multiplications in all layers - the same accuracy that we also use in our implementation. 
Furthermore, the accuracies do not match since we always use the full, uncropped images as input.
The high sparsity in the Human3.6M dataset together with the efficient implementation allows us to accelerate inference significantly.
Compared to DeltaCNN, which does not support varying resolution input or moving cameras, we were able to achieve 13\% higher frame rates with similar accuracy.
We level the playing field in comparisons against the dense cuDNN backend by inferring the same cropped images that we use as input for \ours.
Still, we outperform cuDNN by up to 3.2x, and achieve a speedup of up to 3.2x over dense DeltaCNN with full resolution input.
For Human3.6M, we know that the actors never cover the entire frame at once.
We use this knowledge to lower the size of the buffers by 25\% on each axis - reducing the memory footprint of buffers by 44\% in total.
These savings allow us to increase the batch size even further.
When fully utilizing all available memory with an increased batch size, we achieve frame rates up to 3.9x over the cropped cuDNN baseline and 37\% over DeltaCNN.



\textbf{Ablation Study}
We evaluated the individual impact of our contributions in an ablation study (see Table~\ref{tab:ablation}). 
\begin{table}[h]
    \centering
    \begin{tabular}{l|c|c|c}
                    & FPS & IoU & \# resets \\
\hline
cuDNN               & 40.9 & 79.59 & -   \\
DeltaCNN            & 44.1 & 79.38 & -   \\
\textbf{MotionDeltaCNN} & \textbf{75.1} & \textbf{79.43} & \textbf{8}  \\
- w/o dynamic init  & 56.4 & 79.63 & 110 \\
- w/o padded conv.  & 65.3 & 77.89 & 8  \\
- w/o ROI scaling   & 63.6 & 79.33 & 8  \\
- w/o ROI \& noise* & 50.9 & 79.06 & 8  \\
    \end{tabular}
    \caption{Ablation study. Evaluating BMVOS on DAVIS2017 dataset using a Nvidia GTX 1080 Ti with a batch size of 5. We evaluate on the subset of videos with significant camera motion. * increased threshold on video input.}
    \label{tab:ablation}
\end{table}
\textit{Without dynamic initialization}, \ours always resets the buffers when new image tiles are unveiled.
After a reset, the next frame is processed densely.
This has a negative impact on the speedup, but slightly increases the accuracy as it regularly flushes accumulated errors.
\textit{Padded convolutions} are key in enabling spherical buffers. 
Without them, new tiles are not fused into existing buffers correctly, leading to a reduction in accuracy.
Our evaluation shows that \textit{(ROI) focused sensing} enables great speedups without lowering the accuracy of the segmentation.
By \textit{disabling noise suppression} as well, the model becomes more sensitive to errors in image alignment and small changes in the image. 
To compensate for this, we increase the threshold of the first truncation from $0.15$ to $0.35$ for this evaluation.
However, even with an increased threshold, the speedup is significantly smaller compared to default truncation, while the accuracy is already much lower.
Overall, the ablation study shows that all parts of MotionDeltaCNN work in tandem to achieve the best speedup without large impact on accuracy.
\vspace{-7pt}
\section{Discussion}
\vspace{-3pt}
Our evaluations prove that \ours can accelerate CNN video inference with moving cameras where existing sparse methods fail to outperform the dense reference. 
However, our evaluation also shows that freely moving cameras, especially when combined with dense updates as in the DAVIS dataset, lead to a much smaller speedup than the static camera setup with sparse motion in Human3.6M.
Furthermore, the additional padding slightly increases memory and compute overhead over DeltaCNN.



\textbf{Receptive Field}
While \ours does achieve outputs comparable to standard CNN inference, it should be noted that even in dense mode, the results are not identical.
This can be observed in Figure \ref{fig:dilated_padded_convolution} where the fused output of \ours matches the result using the fused input in row $a$, which includes information from neighbor pixels seen in previous frames.
Since we do not actively propagate inverted accumulated values once a previously processed region gets out of view, the receptive field of later layers, also for pixels inside the current view, still contains some information from that now unseen area.
This can have positive and negative effects alike.
These features might not be valid anymore.
At the same time, some out-of-view features can help to make better predictions due to additional context.

\textbf{Applications}
\ours is a general-purpose framework for video inference and can thus be used with all pre-trained CNNs by replacing the original, dense layers with our sparse equivalent.
As additional input, \ours requires camera extrinsics for the frame to be processed. 
While most publicly available datasets do not include camera extrinsics and therefore require image stabilization, smartphones, tablets and VR headsets nowadays have built-in support for tracking the camera pose between frames in real-time.
Potential applications are not limited to classic computer vision applications like tracking and segmentation, but it could also be used for video up-scaling, filtering, etc.
However, applications like image-denoising (diffusion) are likely not a good fit as they provide low levels of spatial update sparsity between iterations and they are very sensitive to noise. 

\textbf{Limitations}
One of the main limitations of our approach is its dependency on robust video stabilization.
Our experiments with the DAVIS dataset showed that simple video stabilization techniques can break when blur or parallax is involved, or when it fails to find robust features. 
Parallax can greatly reduce the gains of our method, even when the scene itself is mostly static. 
When foreground and background layers move at different speeds, frames can only be properly aligned to stabilize one of these layers - leading to dense updates in layers moving at different speeds.

\ours also comes with the trade-off between buffer overallocation and losing information by cropping the input to match the shape of the buffer.
In our evaluations, when the camera zooms or rotates around the forward axis (\emph{roll}), the aligned input can exceed the size of the buffers, and therefore requires slight cropping to fit into the target shape (see Figure~\ref{fig:results}).
Alternatively, the buffers could also be increased to allow for larger inputs in upcoming frames - at the cost of higher memory consumption.
\vspace{-6pt}
\section{Conclusion}
\vspace{-4pt}
\label{sec:conclusion}
We propose \ours, an extension to the sparse video CNN framework DeltaCNN adding support for moving camera input.
To the best of our knowledge, we are the first to propose a solution for moving cameras with sparse inference.
\ours uses spherical buffers, padded convolutions and dynamic initialization of newly unveiled regions to support combining previously processed pixels and newly unveiled pixels together with high sparsity.
We test our approach in practice by implementing all kernels natively on the GPU to evaluate the real-world performance with modern hardware. 
Our experiments show that we outperform DeltaCNN by up to 90\% on moving camera input and that \ours can also be beneficial to lower the memory footprint in static camera scenarios.

{\small
\bibliographystyle{ieee_fullname}
\bibliography{main}
}

\section*{Supplementary Material}
In this document, we provide further details about the implementation, evaluation and optimizations.

\subsection*{Maximum Pooling}
In the original DeltaCNN implementation, maximum pooling layers process the result for newly accumulated inputs and for previously accumulated inputs, which is then subtracted to obtain the Delta output of the layer.
In \ours, this approach does not work anymore as soon as regions inside the accumulated input values buffer are reset to zero.
When the pooling kernel accesses a pixel's neighboring pixels, with some of them reset and others retained from previous frames, the output for the previous frame cannot be processed correctly anymore - leading to incorrect Delta values.
We solve this using an additional buffer storing the previous output of a pixel.
Since the Delta value is always relative to the previous output, storing it implicitly solves the dependency on neighboring pixel states.
Each pixel was either reset itself in the current frame - in this case, we process the Delta against a zero reference -
or it was retained from previous frames - in which case we process the Delta against the previous output.

\subsection*{Video Object Segmentation Evaluation Details}
In video object segmentation, we reduce the update rate by scaling updates outside the region of interest, i.e., the segmentation mask of the previous frame.
We dilate the previous mask multiple times using maximum pooling with different kernel sizes (10, 20, 40) and compute the average pixel values of the three masks $m$.
The image Delta is then scaled using a factor of $0.4 + 0.6 * m$ before comparing it against the truncation threshold. 

The truncation thresholds for BMVOS are auto-tuned on a sequence from the training dataset (tractor-sand) using a threshold of 0.02 as a starting point for all layers but the first. 
The first layer uses 0.15 and an update mask dilation of 10 pixels (a single pixel update would be dilated over 21x21 pixels).
We do not use the same, sensitive thresholds for DeltaCNN since they lead to dense updates nearly every frame due to the lack of image alignment.
Instead, we increase the threshold of the first layer to 0.2 and auto tune the remaining thresholds on the same sequence without frame alignment.
This leads to a slightly lower accuracy for DeltaCNN than for \ours.
We deliberately target a lower accuracy when tuning the thresholds for DeltaCNN since we aim to show that we outperform DeltaCNN even when they are allowed to truncate updates more aggressively.
The auto-tuning procedure is implemented as an iterative front-to-back process as described in DeltaCNN \cite{Parger_2022}.

\subsection*{Further Optimizations}

\paragraph{Implicit bias application}
Adding biases onto newly allocated tiles requires launching a separate kernel on the GPU after every convolutional and batch normalization layer if they use biases.
However, layers that are directly followed by activation \& truncation layers can skip this step and add the bias implicitly during activation.
By initializing the newly allocated tiles of the activation layer's truncated values buffer with the bias instead setting it to zero, the bias is automatically added when applying the activation. 
This way, we can reduce the overhead of attaching new tiles to existing buffers.

\paragraph{Optimizing memory consumption}
The memory overhead of \ours can be a limiting factor on low-end hardware.
But in many cases, memory consumption can be reduced by disabling truncation on selected activation layers at the cost of increased sparsity.
For example, the DenseNet \cite{Huang2017} backbone of BMVOS \cite{Cho2022} consists mainly of pairs of 3x3 and 1x1 convolutions.
In this network, we only truncate in the activation layer following a 3x3 convolution in which updates are dilated.
This way, every second activation layer needs only an accumulated values buffer - saving 50\% of the layer's memory at the cost of slightly higher update rates.

\subsection*{Qualitative Results Human Pose Estimation}
\begin{figure*}[pt]
\centering
\begin{subfigure}{.05\linewidth}
    Dense
\end{subfigure}%
\begin{subfigure}{.130\linewidth}
  \centering
  \includegraphics[width=.95\linewidth,valign=c,trim={170px 150px 10px 30px},clip]{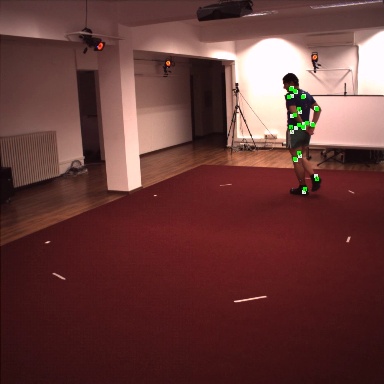}
\end{subfigure}%
\begin{subfigure}{.130\linewidth}
  \centering
  \includegraphics[width=.95\linewidth,valign=c,trim={170px 150px 10px 30px},clip]{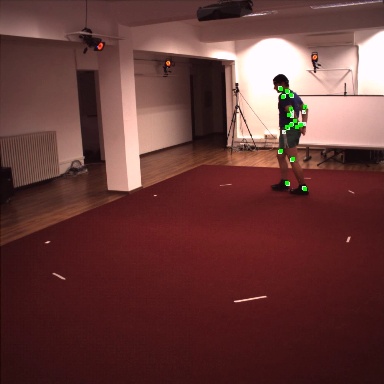}
\end{subfigure}%
\begin{subfigure}{.130\linewidth}
  \centering
  \includegraphics[width=.95\linewidth,valign=c,trim={170px 150px 10px 30px},clip]{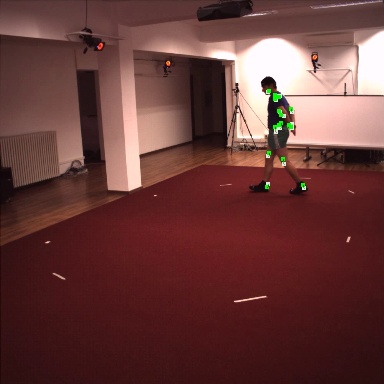}
\end{subfigure}%
\begin{subfigure}{.130\linewidth}
  \centering
  \includegraphics[width=.95\linewidth,valign=c,trim={170px 150px 10px 30px},clip]{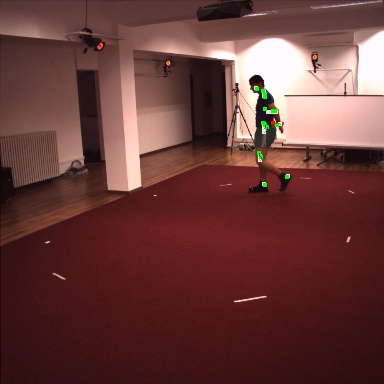}
\end{subfigure}%
\begin{subfigure}{.130\linewidth}
  \centering
  \includegraphics[width=.95\linewidth,valign=c,trim={170px 150px 10px 30px},clip]{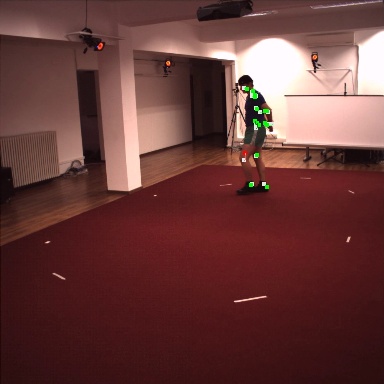}
\end{subfigure}%
\begin{subfigure}{.130\linewidth}
  \centering
  \includegraphics[width=.95\linewidth,valign=c,trim={170px 150px 10px 30px},clip]{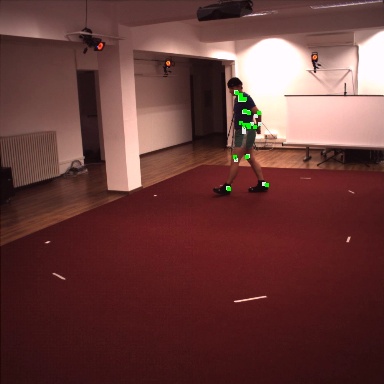}
\end{subfigure}%
\begin{subfigure}{.130\linewidth}
  \centering
  \includegraphics[width=.95\linewidth,valign=c,trim={170px 150px 10px 30px},clip]{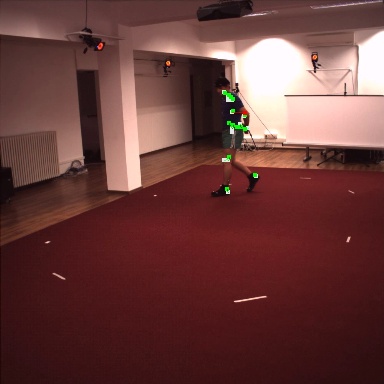}
\end{subfigure}

\begin{subfigure}{.05\linewidth}
    DC
\end{subfigure}%
\begin{subfigure}{.130\linewidth}
  \centering
  \includegraphics[width=.95\linewidth,valign=c,trim={170px 150px 10px 30px},clip]{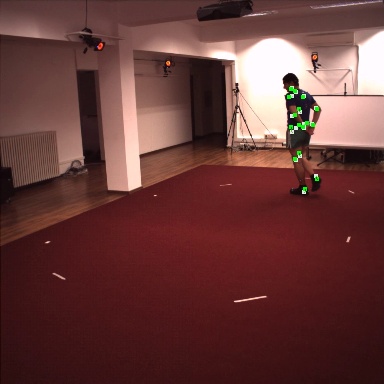}
\end{subfigure}%
\begin{subfigure}{.130\linewidth}
  \centering
  \includegraphics[width=.95\linewidth,valign=c,trim={170px 150px 10px 30px},clip]{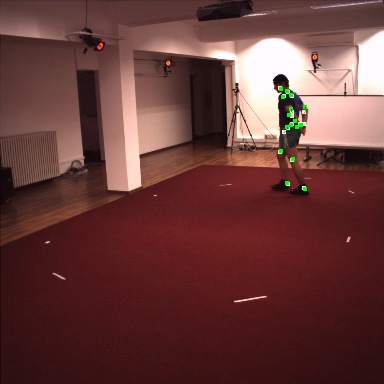}
\end{subfigure}%
\begin{subfigure}{.130\linewidth}
  \centering
  \includegraphics[width=.95\linewidth,valign=c,trim={170px 150px 10px 30px},clip]{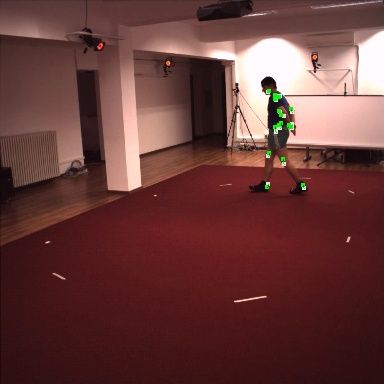}
\end{subfigure}%
\begin{subfigure}{.130\linewidth}
  \centering
  \includegraphics[width=.95\linewidth,valign=c,trim={170px 150px 10px 30px},clip]{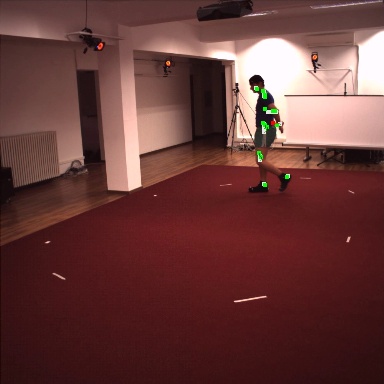}
\end{subfigure}%
\begin{subfigure}{.130\linewidth}
  \centering
  \includegraphics[width=.95\linewidth,valign=c,trim={170px 150px 10px 30px},clip]{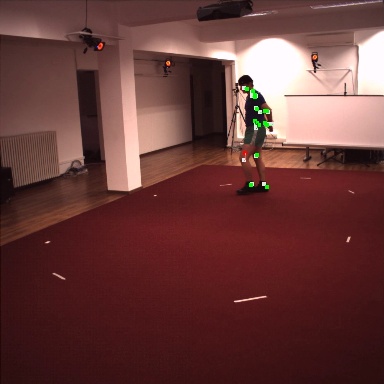}
\end{subfigure}%
\begin{subfigure}{.130\linewidth}
  \centering
  \includegraphics[width=.95\linewidth,valign=c,trim={170px 150px 10px 30px},clip]{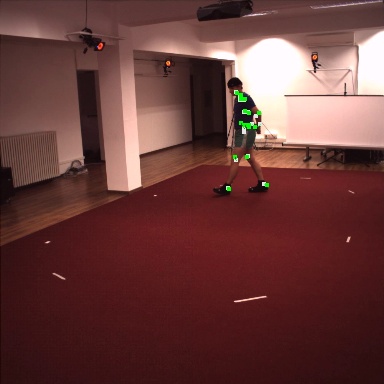}
\end{subfigure}%
\begin{subfigure}{.130\linewidth}
  \centering
  \includegraphics[width=.95\linewidth,valign=c,trim={170px 150px 10px 30px},clip]{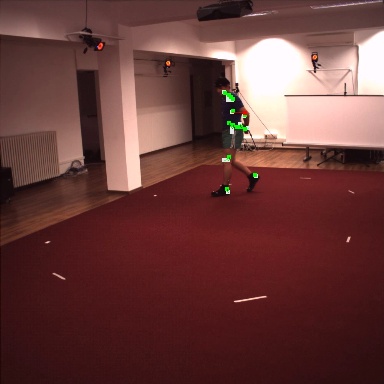}
\end{subfigure}

\begin{subfigure}{.05\linewidth}
    Ours
\end{subfigure}%
\begin{subfigure}{.130\linewidth}
  \centering
  \includegraphics[width=.95\linewidth,valign=c,trim={170px 150px 10px 30px},clip]{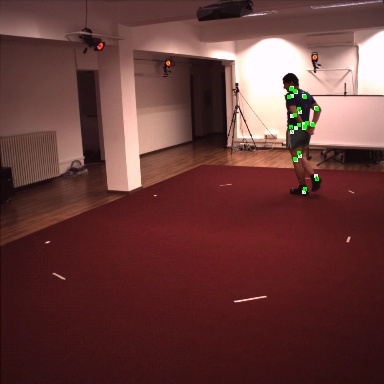}
\end{subfigure}%
\begin{subfigure}{.130\linewidth}
  \centering
  \includegraphics[width=.95\linewidth,valign=c,trim={170px 150px 10px 30px},clip]{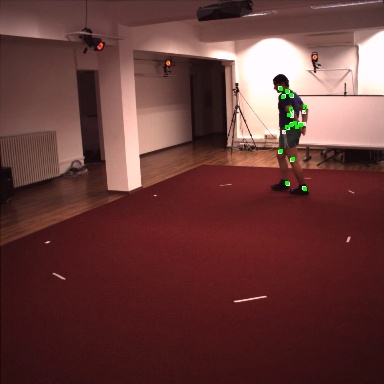}
\end{subfigure}%
\begin{subfigure}{.130\linewidth}
  \centering
  \includegraphics[width=.95\linewidth,valign=c,trim={170px 150px 10px 30px},clip]{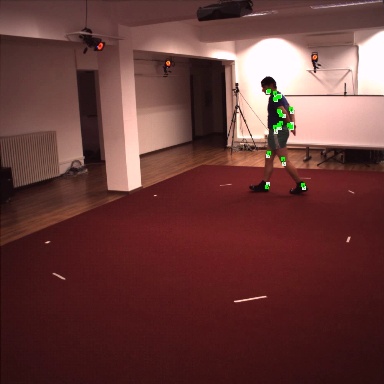}
\end{subfigure}%
\begin{subfigure}{.130\linewidth}
  \centering
  \includegraphics[width=.95\linewidth,valign=c,trim={170px 150px 10px 30px},clip]{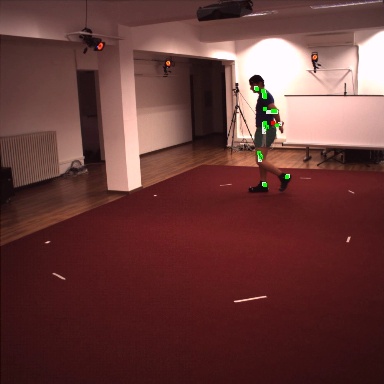}
\end{subfigure}%
\begin{subfigure}{.130\linewidth}
  \centering
  \includegraphics[width=.95\linewidth,valign=c,trim={170px 150px 10px 30px},clip]{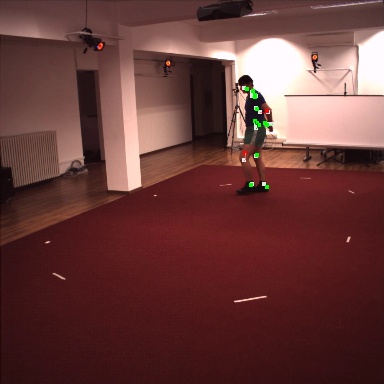}
\end{subfigure}%
\begin{subfigure}{.130\linewidth}
  \centering
  \includegraphics[width=.95\linewidth,valign=c,trim={170px 150px 10px 30px},clip]{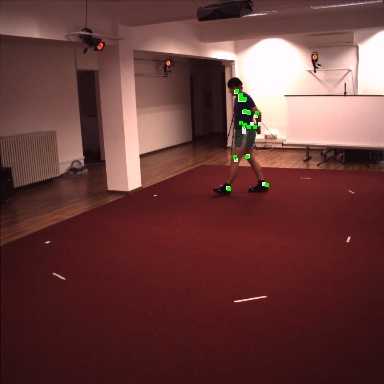}
\end{subfigure}%
\begin{subfigure}{.130\linewidth}
  \centering
  \includegraphics[width=.95\linewidth,valign=c,trim={170px 150px 10px 30px},clip]{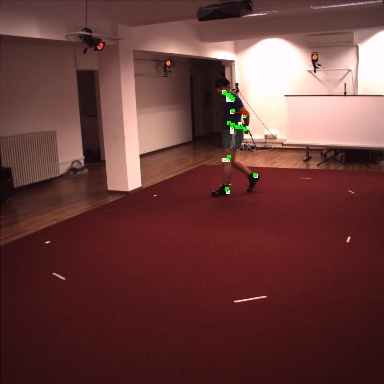}
\end{subfigure}

\begin{subfigure}{.05\linewidth}
  \centering
  \caption*{}
\end{subfigure}%
\begin{subfigure}{.130\linewidth}
  \centering
  \caption*{Frame 0}
\end{subfigure}%
\begin{subfigure}{.130\linewidth}
  \centering
  \caption*{Frame 10}
\end{subfigure}%
\begin{subfigure}{.130\linewidth}
  \centering
  \caption*{Frame 20}
\end{subfigure}%
\begin{subfigure}{.130\linewidth}
  \centering
  \caption*{Frame 30}
\end{subfigure}%
\begin{subfigure}{.130\linewidth}
  \centering
  \caption*{Frame 40}
\end{subfigure}%
\begin{subfigure}{.130\linewidth}
  \centering
  \caption*{Frame 50}
\end{subfigure}%
\begin{subfigure}{.130\linewidth}
  \centering
  \caption*{Frame 60}
\end{subfigure}

\caption{
Qualitative comparison between conventional, dense inference, DeltaCNN (DC) and \ours using HRNet on the Human3.6M dataset.
For better visibility, we zoomed in closer to the subject (2x zoom).
White markers are the ground truth joint positions. 
Green and red are the predicted joint positions, with red indicating an incorrect prediction according to PCKh@0.5.
}
\label{fig:h36m_qualitative}
\end{figure*}
Figure~\ref{fig:h36m_qualitative} displays a qualitative comparison between the joint positions predicted using the dense reference, DeltaCNN and \ours.
As can be seen in this example, the predicted joint locations are identical in a all 3 cases for the majority of the joints.
However, in some cases, small differences between the approaches can be noticed.
The most significant being in Frame 40, where \ours predicts two joints incorrectly, whereas DeltaCNN and the dense reference each only have one incorrect prediction.
At closer inspection, it can be seen that the predicted joint position of the elbow only varies slightly between the dense reference and MotionDeltaCNN, just enough to exceed the accuracy threshold.

\end{document}